\RequirePackage{fix-cm}
\documentclass[smallcondensed]{svjour3}       % onecolumn (second format)
\usepackage{siunitx}

\usepackage{float}
\usepackage{verbatim}
\usepackage{svg}

\usepackage[T1]{fontenc}
\usepackage{times}%

\usepackage[square,numbers]{natbib}
\usepackage{hyperref}
\usepackage{dcolumn}
\usepackage{graphics}

\usepackage{blindtext}
\usepackage{graphicx}
\usepackage{tikz}

\usepackage[T1]{fontenc}
\usepackage[utf8]{inputenc}
\usepackage{hyphenat}
\usepackage[ruled,vlined]{algorithm2e} % conflicts with style 'iosart2c', since algorithm is already defined
\usepackage[noend]{algpseudocode}
\usepackage{booktabs}
\usepackage{placeins}
\usepackage{array}
\usepackage{multirow}
\usepackage{amsmath,amssymb}
\usepackage{calc}
\usepackage{xcolor}
\usepackage{epstopdf}
\usepackage{threeparttable}
\usepackage{caption}
\usepackage{subcaption}

\newcommand{\labelix}{j}

\newcommand{\xx}{{\pmb{x}}}
\newcommand{\WW}{{\pmb{W}}}
\newcommand{\zz}{{\pmb{s}}}
\newcommand{\yy}{{\pmb{y}}}
\newcommand{\hh}{\mathcal{H}}
\newcommand{\LL}{\mathcal{L}}

\newcommand{\TODO}[1]{\vspace{0.01cm}\textcolor{red}{TO DO:\ #1}\vspace{0.01cm}}

\newcommand{\refeq}[1]{equation~(\ref{eq:#1})}
\newcommand{\reffig}[1]{fig.~\ref{fig:#1}}
\newcommand{\refFig}[1]{Fig.~\ref{fig:#1}}
\newcommand{\reftab}[1]{table~\ref{tab:#1}}
\newcommand{\refTab}[1]{Table~\ref{tab:#1}}

\newcommand{\refSec}[1]{Section~\ref{sec:#1}}
\newcommand{\refalg}[1]{algorithm~\ref{algorithm:#1}}
\newcommand{\refAlg}[1]{Algorithm~\ref{algorithm:#1}}

\newif\ifREVISAO
\REVISAOtrue
\newif\ifDIFF
\DIFFtrue
\ifREVISAO
    \newcommand{\removido}[1]{\textcolor{red}{#1}}
    \newcommand{\flavio}[1]{\textcolor{orange}{#1}}
    \newcommand{\victor}[1]{\textcolor{cyan}{#1}}

\else
    \ifDIFF
        \newcommand{\removido}[1]{}
        \newcommand{\flavio}[1]{}
        \newcommand{\victor}[1]{}
        \newcommand{\TODO}[1]{}
    \else
        \newcommand{\removido}[1]{\textcolor{red}{\iffalse #1 \fi}}
        \newcommand{\flavio}[1]{\textcolor{black}{#1}}
        \newcommand{\victor}[1]{\textcolor{black}{#1}}
    \fi
\fi

\newcommand{\RAKEL}[0]{RA\emph{k}EL}

\newcolumntype{L}[1]{>{\raggedright\let\newline\\\arraybackslash\hspace{0pt}}m{#1}}
\newcolumntype{C}[1]{>{\centering\let\newline\\\arraybackslash\hspace{0pt}}m{#1}}
\newcolumntype{R}[1]{>{\raggedleft\let\newline\\\arraybackslash\hspace{0pt}}m{#1}}
\newcolumntype{H}{>{\setbox0=\hbox\bgroup}c<{\egroup}@{}}

\begin{document}
%\begin{frontmatter}			      % The preamble begins here.

\title{Exploring label correlations using decision templates for ensemble of classifier chains}



\author{Victor F. Rocha*       \and Alexandre L. Rodrigues \and Thiago Oliveira-Santos
  \and        Fl\'{a}vio M. Varej\~{a}o 
}


\institute{Victor F. Rocha* \at
              Department of Computer Science, Federal University of Esp\'{\i}rito Santo, Vit\'{o}ria, Esp\'{\i}rito Santo, 29075-910, Brazil \\ 
              \email{victor.rocha@ufes.br}               %  \\
            \and
           Alexandre L. Rodrigues \at
              Department of Statistics, Federal University of Esp\'{\i}rito Santo, Vit\'{o}ria, Esp\'{\i}rito Santo, 29075-910, Brazil \\
              \email{alexandre.rodrigues@ufes.br}
            \and
           Thiago Oliveira-Santos \at
              Department of Computer Science, Federal University of Esp\'{\i}rito Santo, Vit\'{o}ria, Esp\'{\i}rito Santo, 29075-910, Brazil \\
              \email{todsantos@inf.ufes.br}
            \and
           Fl\'{a}vio M. Varej\~{a}o \at
              Department of Computer Science, Federal University of Esp\'{\i}rito Santo, Vit\'{o}ria, Esp\'{\i}rito Santo, 29075-910, Brazil \\
              \email{fvarejao@inf.ufes.br}
            \and
            *Corresponding author: Victor F. Rocha, \email{victor.rocha@ufes.br}
}

\date{Submitted on 13 Mar 2026}

\maketitle

\begin{abstract}
The use of ensemble-based multi-label methods has been shown to be effective in improving multi-label classification results.
One of the most widely used ensemble-based multi-label classifiers is Ensemble of Classifier Chains.
Decision templates for Ensemble of Classifier Chains (DTECC) is a fusion scheme based on Decision Templates that combines the predictions of Ensemble of Classifier Chains using information from the decision profile for each label, without considering information about other labels that might contribute to the classified result. 

Based on DTECC, this work proposes the Unconditionally Dependent Decision Templates for Ensemble of Classifier Chains (UDDTECC) method, a classifier fusion method that seeks to exploit correlations between labels in the fusion process. In this way, the classification of each label in the problem takes into account the label values that are considered conditionally dependent and that can lead to an improvement in the classification performance.

The proposed method is experimentally compared with two traditional classifier fusion strategies and with a stacking-based strategy.
Empirical evidence shows that using the proposed Decision Templates adaptation can improve the performance compared to the traditionally used fusion schemes on most of the evaluated metrics.

% Please include a maximum of seven keywords
\end{abstract}

%\end{frontmatter}

\section{Introduction} \label{sec:introduction}

In multi-label classification, an example may belong to multiple classes at the same time. Many real world problems can be better represented as multi-label, for example, the prediction of subcellular locations of proteins based on their sequences \cite{xu2016multi} and the classification of music according to the emotion that it evokes on the listener \cite{tsoumakas2008effective}.

Ensembles of classifiers are defined as groups of classifiers trained on the same classification problem that work together to produce a robust result. Since each classifier produces its own result, these individual results must be combined via a fusion function to produce the final classification result. The use of ensembles has been shown to be an effective technique for improving performance in multi-label classification.

Two main strategies are used for the multi-label task: developing methods specifically adapted for the multi-label problem (adapted algorithms), and transforming the multi-label task into a set of traditional single-label classification tasks (problem transformation methods) whose results are combined to obtain a multi-label result.
Problem transformation has proven to be a popular strategy, as it allows the use of the wide range of single-label methods that are already available and well established in the literature for multi-label tasks.

Among the many ensemble-based multi-label transformation methods proposed is the Ensemble of Classifier Chains (ECC) \cite{read2011classifier}, a method that excels in classification performance when compared to other commonly used multi-label methods \cite{moyano2018review}. 
To enable the use of an ensemble of classifiers as an alternative to classification, it is necessary to combine the predictions of each individual classifier in a final prediction that is ideally more robust. For this purpose, several methods have been proposed, among them the Majority Vote (MV) and Mean Ensemble (ME) methods. The DTECC  method, a version of the Decision Templates (DT) \cite{kuncheva2001decision}  method applied in a multi-label context and specifically adapted for fusing classifiers from an ECC, has proven effective mainly with respect to the Accuracy and F-Measure metrics \cite{freitas2022ensemble}.

In this paper, a new strategy based on the DTECC method is proposed, evaluated and compared with traditional fusion methods and a stacking-based method. This variation, called Unconditionally Dependent Decision Templates for Ensemble of Classifier Chains (UDDTECC), is based on the idea of exploring possible correlations between labels, using in the classification of each label all the support information regarding the labels that are found to be unconditionally dependent on that label.

The $\phi$ coefficient, which aims to measure the degree of association between two binary variables, \cite{ekstrom2011phi, kemp2003applied} is used to identify the existing unconditional correlations between labels. In the classification of a label, only the support values for the labels that are judged to be unconditionally dependent on that label will be present.
This new method is compared to the traditional MV and ME fusion schemes often used in conjuntion with ECC and to the original DTECC. 
The proposed DT-based scheme can be compared to the use of a meta-classifier as it occurs in stacking methods, therefore an appropriate stacking-based method is also included in the experiments and compared directly with UDDTECC.

The paper is structured as follows: \refSec{background} summarizes the main theoretical foundations of the concepts about multi-label, ensemble of classifiers, fusion functions and the ECC classifier. \refSec{DTECC} 
reviews the DTECC method and \refSec{UDDTECC}
shows the new UDDTECC method, one variation of the DTECC method.  \refSec{setup} presents and discusses the experimental results. Finally, \refSec{conclusion} highlights some important findings of this work.

\section{Background}
\label{sec:background}
In this section, the main theoretical foundations necessary for the comprehension of the DTECC method and its variations are presented. 

\subsection{Multi-label classification} \label{sec:ml_ensembles}
Multi-label classification, a generalization of the single-label classification problem, allows an observation (or instance) to be associated with multiple labels. This generalization allows us to more accurately model many real-world problems.
Let $L = \left\{\ell_{1},\ell_{2},\ell_{3} \ldots, \ell_{m} \right\}$ be the set of possible labels of a multi-label classification problem with $|L| = m$, all $2^{|L|}$  unique label combinations are therefore acceptable answers. It becomes critical to understand the correlations between the labels in order to extract the best possible result \cite{zhu2017multi}. A multi-label classifier can be defined as a mapping $\pmb{h}: \mathcal{X} \rightarrow \mathcal{Y}$, $\xx \mapsto \yy$
of the feature space
$\mathcal{X}$ for the class domain $\mathcal{Y} = \{0,1\}^{m}$.
A multi-label dataset is defined as $n$ pairs of feature vectors and
binary vectors of dimension $m$ as
$S = \left\{(\xx^{1},\yy^{1}), \ldots, (\xx^{n},\yy^{n}) \right\}$, with
$\yy^{k} = \{y^{k}_{1}, \ldots,y^{k}_{m} \} $, $y^{k}_{j} \in \{0,1\}$, $k \in [1,n]$, $j \in [1,m]$, where $y^{k}_{j} = 1$ represents the presence of the label $\ell_j$ in example $k$ and $y^{k}_{j} = 0$ represents its absence. 

\subsection{Classifiers Ensemble} 

The technique of combining the predictions of multiple classifiers to produce a single classifier is known as ensemble of classifiers \cite{optiz1999ensemble, rokach2010ensemble}. 
Ideally, a classifier ensemble should achieve better performance than any of the individual classifiers of the ensemble. Good ensembles are generally composed by individual  classifiers which are both accurate and make their errors in different parts of the feature space.
Ensemble methods can be divided into two categories \cite{kuncheva2001decision, woods1997combination, kuncheva2002switching, ho2000complexity}: classifier selection  and classifier fusion. 

Classifier selection methods train classifiers to be local experts in some area of the feature space. The importance given to the prediction of a specific classifier is inversely proportional to the distance of the data used to train this classifier. One or more classifiers can be selected to give the final decision \cite{woods1997combination, jacobs1991adaptive,  alpaydin1996local, giacinto2001approach}.

In classifier fusion, classifiers are trained on the entire feature space. In those techniques the output of the weaker classifiers must be merged to create the stronger output. Traditional methods such as Bagging \cite{breiman1996bagging} and Boosting \cite{schapire1990strength, freund1999short} are based on this approach.
These methods are based on resampling techniques for obtaining different training sets for each of the individual classifiers. 
This paper focuses on classifier fusion methods. 
Thus, further explanation on the operation of classifier fusion methods is presented.

\subsection{Single-label Fusion Functions}

Let $\mathcal{H} = \{ \hat{h}_1,\ldots,\hat{h}_c \}$ be the set of classifiers of a classifier ensemble. The output of each classifier can be scaled to the [0,1] interval without loss of generality. 

For a feature vector $\xx \in \mathcal{X}$, the output of the \textit{i}th classifier is represented as 
$\hat{h}_i(\xx)=[d_{i,\ell_{1}}(\xx),\ldots,d_{i,\ell_{m}}(\xx)], d_{i,\ell_{j}}(\xx) \in [0,1], 1 \leq j \leq m $,
where $d_{i,\ell_{j}}(\xx)$ can be generally interpreted as the degree of \textit{support} outputted by the $\hat{h}_i$ classifier to the hypothesis that $\xx$ is labeled as $\ell_j$. The output of all the classifiers that compose an ensemble composed by $c$ classifiers and $m$ labels can be organized as a matrix known as the Decision Profile (DP) of $\xx$:

\begin{equation*}
\text{DP}(\xx)=\left(
\begin{array}{ccccc}
d_{1,\ell_1}(\xx) & \cdots & d_{1,\ell_j}(\xx) & \cdots & d_{1,\ell_m}(\xx)  \\
\cdots & & & & \\
d_{i,\ell_1}(\xx) & \cdots & d_{i,\ell_j}(\xx) & \cdots & d_{i,\ell_m}(\xx)  \\
\cdots & & & & \\
d_{c,\ell_1}(\xx) & \cdots & d_{c,\ell_j}(\xx) & \cdots & d_{c,\ell_m}(\xx)  \\
\end{array}
\right).
\label{eq:DP}
\end{equation*}

The output $\hat{h}(\xx)$ of the classifier $\mathcal{H}$ is defined by a \textit{fusion function} $\mathcal{F}$ that uses the classifiers outputs $\hat{h}_i(\xx)$ to create a combined output:

\begin{equation}
\hat{h}(\xx) = \mathcal{F}(\hat{h}_1(\xx),\cdots,\hat{h}_c(\xx)) = \mathcal{F}(\text{DP}(\xx)).
\end{equation}

Fusion functions can be divided into two groups based on whether they require additional training once the classifiers of the ensemble are trained \cite{wang2021active}. Popular combination functions that do not require further training include the majority vote, maximum, minimum, average, and product. 

Except for the majority vote, which uses DP($\xx$) for computing the label that is most voted by the classifiers, all these functions compute the combined support value for each label as 
\begin{equation}
    d_j (\xx) = f(d_{1,\ell_j}(\xx),\ldots,d_{c,\ell_j}(\xx)),
\end{equation}
where $f$ represents the corresponding mathematical function (maximum, minimum, average, or product). For single-label classification, the resulting label is calculated using \refeq{harden}: 

\begin{equation}
\hat{h}(\xx)=\ell \Leftrightarrow d_{\ell}(\xx) = \max_{j=1,\ldots,m} \{d_j(\xx)\}.
\label{eq:harden}
\end{equation}

\subsection{Decision Templates} \label{decisiontemplates}

Decision Templates \cite{kuncheva2001decision} are an example of a fusion function that require additional training. Let $S = \left\{ {\zz}_{1}, \ldots, {\zz}_{n} \right\}$ be the single-label training set. An ensemble of classifiers is trained using $S$. For each label $l_j \in L$ the following matrix is computed:

\begin{equation}
\text{DT}_j = \frac{ \sum _{k=1}^{n} f(\zz_k,j) \cdot \text{DP}(\zz_k) }{ \sum _{k=1}^{n} f(\zz_k,j)},
\label{eq:DT}
\end{equation}
where $f(\zz_k,j)$ is a function that assumes the value 1 if the label of $\zz_k$ is $l_j$. Otherwise, the function is 0. 

$\text{DT}_j$ can be defined as the expected value of $\text{DP}(\xx)$ for the $l_j$ label. The final value of support to $l_j$ is directly proportional to the level of \textit{similarity} between the matrices $\text{DP}($\xx$)$ and $\text{DT}_j$. 

The proximity function based on the Euclidean distance recommended in \cite{rogova1994combining} is calculated as: 
\begin{align}
\theta_{j}(\xx) = \frac{1 - ||\text{DP}(\xx)-\text{DT}_{j}||_{2}^{2}}{ \sum_{l=1}^{m}(1 - ||\text{DP}(\xx)-\text{DT}_{l}||_{2}^{2}) }.
\label{eq:similar1994}
\end{align}

This work adopts a similarity function in the fusion process based on \refeq{similar1994} but removes the normalization:
\begin{align}
\mu_{j}(\xx) = 1 - ||\text{DP}(\xx)-\text{DT}_{j}||_{2}^{2}= \nonumber \\
1 - \sum_{i=1}^{c} \sum_{l=1}^{m}
	\left[ d_{i,l}(\xx) - (\text{DT}_{j})_{i,l} \right]^{2},
\label{eq:similar}
\end{align}

\noindent where $||.||_{2}$ is the entry-wise matrix norm, or equivalently
the Euclidean distance between the linearized DP and DT (all lines
have been sequentially assembled into a $c\times m$ vector).
After the support $\mu_{j}$ has been obtained for each class,
a further evaluation takes the final decision. Usually, in a
single-label crisp classification
task, the highest score defines the assigned class.
In a binary classification task, the first class is chosen
if $\mu_{1}(\xx)\geq \mu_{2}(\xx)$.
In a multi-label setting, this strategy is not applicable. 

\subsection{Ensemble of Classifier Chains} \label{multi-labelensemble}

 The multi-label classification method Classifier Chains (CC) \cite{read2011classifier} is categorized as a problem transformation method, i.e., it transforms the multi-label problem into $m$ single-label problems to allow the use of single-label methods.

To explore possible correlations between the labels, the CC method first defines the order in which the problem labels will be evaluated and for each evaluated label it incrementally adds the classification results using the evaluated labels. The first classifier of the string $\hat{h}_{1}^{\mathrm{CC}}: \mathcal{X} \rightarrow \{0,1\}$, $\xx \mapsto y_{1}$ works as one of the $ m $ single-label binary classifiers. The prediction value of the first label is used in conjunction with the feature vector $ \xx $ to perform the prediction of the second label.
$\hat{h}_{2}^{\mathrm{CC}}: \mathcal{X} \times \{0,1\} \rightarrow \{0,1\}$,
$(\xx, y_{1}) \mapsto y_{2}$.
The classifiers that compose the CC are trained using the real labels $y_{1}$, whereas in the prediction phase the estimated label values $\hat{y}_1$ are used in the mapping, such that $ (\xx, \hat{y}_1) \mapsto \hat{y}_2$.
Therefore, CC is composed of $ m $ classifiers that can use from zero to $ (m-1) $ labels as additional attributes, such that in the training phase $\hat{h}_{j}^{\mathrm{CC}}: \mathcal{X} \times \{0,1\}^{j-1} \rightarrow \{0,1\}$,
$(\xx, y_{1},\ldots,y_{j-1}) \mapsto y_{j}, j=1,\ldots,m$ and in the prediction phase $\hat{h}_{j}^{\mathrm{CC}}: \mathcal{X} \times \{0,1\}^{j-1} \rightarrow \{0,1\}$,
$(\xx, \hat{y}_{1},\ldots,\hat{y}_{j-1}) \mapsto \hat{y_{j}}, j=1,\ldots,m$.

The order in which the chaining of results is done has an impact on the final result, and it is therefore an important issue when using the CC method. With this in mind, the ECC method was proposed. In this method $c$ CC classifiers $C_1, \dots, C_c$ are trained. In order to create a diverse set of classifiers, each of these classifiers is trained using a random ordering of labels and a random subspace of the training dataset, allowing each CC model to be unique and capable of making distinct predictions \cite{read2011classifier}.

Classifications performed for each  classifier member of the ECC ensemble are combined by using a multi-label fusion function. 
The next three subsections describe three multi-label fusion functions further used in the experiments.

\subsection{Multi-label Fusion Functions}

Given the nature of the multi-label problem where multiple labels can occur simultaneously, it is necessary to use fusion functions specifically designed for this type of problem. This section presents some fusion methods used for multi-label classifier fusion, more specifically for the ECC method.

\subsubsection{Majority Vote ECC}

The fusion function originally proposed by the authors of the ECC method was the Majority Vote (MV). MV is one of the most straightforward and widely used fusion schemes in problems involving ensembles of multi-label classifiers. In this scheme, the label $\ell_j$ is only assigned to a test example $\xx$ if a certain percentage of set member classifiers (a predefined threshold value) have predicted this label. 
The individual hard label predictions of each classifier are represented as vectors $\yy_i = (y_{i,\ell_1}, \ldots, y_{i,\ell_m})$, where each component $y_{i,\ell_j} \in {0, 1}$. These predictions are aggregated into a sum vector $\WW = (\lambda_1, \ldots, \lambda_m)$, with each component defined as $\lambda_j = \sum^{c}_{i=1} y_{i,\ell_j}$. This sum vector $\WW$ is then normalized to produce a vector $\WW^{norm}$, where each component of $\WW^{norm}$ lies within the range [0, 1], effectively representing a probability distribution. The final classification result $\yy$ is defined by a threshold value $t$ so that $\ell_j \in \yy$ only if $\lambda_j \geq t$. The threshold value is usually set to $50\%$, meaning that for the label $\ell_j$ to be in the final prediction at least half of the classifiers must have predicted this label. This approach is exemplified with an example from the Image data set in \refTab{fusionmajt}. From this point on the combination of the ECC method with the MV fusion scheme is called MVECC.  

\begin{table}[h]
\centering
\begin{tabular}{|l|lllllll|}
\hline
$\LL$ =        & \{ & Desert   & Mountains   & Sea   & Sunset   & Trees   & \} \\ \hline
$\yy_1$  & (  & 1   & 0   & 0   & 1   & 1   & )  \\
$\yy_2$  & (  & 1   & 1   & 0   & 1   & 0   & )  \\
$\yy_3$  & (  & 0   & 0   & 0   & 1   & 0   & )  \\
$\yy_4$  & (  & 1   & 0   & 0   & 1   & 0   & )  \\
$\yy_5$  & (  & 0   & 1   & 0   & 1   & 1   & )  \\ \hline
$\WW        $  & (  & 3   & 2   & 0   & 5   & 2   & )  \\
$\WW_{avg} $  & (  & .60 & .40 & .00 & 1.00 & .40 & )  \\ \hline
$\yy$      & (  & 1   & 0   & 0   & 1   & 0   & )  \\ \hline
\end{tabular}

\caption{Example of Majority Vote  fusion function for classifying an instance of the Image data set with an ensemble consisting of 5 multi-label classifiers. Each instance of the data set represents an image that can be classified according to five possible labels: Desert, Mountains, Sea, Sunset and Trees.
If the average of the binary decisions for each label is larger than the threshold, the label is chosen. In this example, a user defined value of threshold $\emph{t} = 0.5$ defines the labels present in the final classification result. The label set ultimately assigned to the image is Desert + Sunset.} 
\label{tab:fusionmajt}
\end{table} 

\subsubsection{Mean Ensemble ECC}

Similarly to MV, Mean Ensemble (ME) is a popular fusion method based on a simple and straightforward premise to perform its fusion process. ME defines whether or not a label $\ell_\labelix$ is present in the final result of the classification by averaging the support value of all classifiers $\pmb{h} \in \mathcal{H} $ for each label:

\begin{equation}
    d_j (\xx) = \text{AVERAGE}(d_{1,\ell_j}(\xx),\ldots,d_{c,\ell_j}(\xx)),
\end{equation}
the resulting label set $\yy$ is defined by a threshold value $t$ (typically $t = 0.5$) such that $\ell_j \in \yy$ if $d_j \geq t$. 

This scheme is exemplified in \refTab{fusionexample} for the Image data set. The support values for each label are averaged and the final decision is taken. In this example, a user defined value of threshold $\emph{t} = 0.5$ defines the labels present in the final classification result. The resulting label set for this instance is \{Desert, Mountains, Sunset\}.

From this point on, the use of the ME fusion method in conjunction with ECC will be called MEECC.

\begin{table}[H]
\centering
\begin{tabular}{|l|ccccccc|}
\hline
$\mathcal{H}$ =        & \{ & Desert   & Mountains   & Sea   & Sunset   & Trees   & \} \\ \hline
$\pmb{\hat{h}}_1$  & (  & 0.8 & 0.3 & 0.1 & 0.6 & 0.7 & )  \\
$\pmb{\hat{h}}_2$  & (  & 0.7 & 0.8 & 0.1 & 0.8 & 0.3 & )  \\
$\pmb{\hat{h}}_3$  & (  & 0.2 & 0.4 & 0.2 & 0.8 & 0.2 & )  \\
$\pmb{\hat{h}}_4$  & (  & 0.9 & 0.2 & 0.1 & 0.9 & 0.4 & )  \\
$\pmb{\hat{h}}_5$  & (  & 0.3 & 0.8 & 0.2 & 0.8 & 0.8 & )  \\ \hline
$\WW        $  & (  & 2.9 & 2.5 & 0.7 & 3.9 & 2.4 & )  \\
$\WW_{avg} $  & (  & 0.58 & 0.5 & 0.14 & 0.78 & 0.48 & )  \\ \hline
$\yy$      & (  & 1 & 1 & 0 & 1 & 0 & )  \\ \hline
\end{tabular}

\caption{Example of MeanEnsemble fusion function for classifying an instance of the Image data set with an ensemble consisting of 5 multi-label classifiers.}
\label{tab:fusionexample}
\end{table} 

\subsubsection{Stacking ECC}
\label{sec-Stack}

Stacking consists of the technique of combining the classification results of each classifier of the ensemble through a meta-classifier. The ensemble member classifiers, also called first-level classifiers, are trained using the training set and the meta-classifier is trained using the results of the first-level classifiers on the training set. Overall, the stacking ensemble approach is a promising strategy for improving the performance of multi-label classification models.

The stacking method involves two stages: training and testing. During training, the base classifiers are trained on the input data, and their predictions are then used to train the meta-classifier. \refFig{stackecc_dg} depicts the training process of the Stack method, which is further elaborated in \refalg{stack}. In the testing stage, the input data is first used to generate predictions from the base classifiers, and then these predictions are used as input to the meta-classifier to produce the final prediction. \refAlg{classifystack} illustrates the classification of an unseen instance, which is exemplified in \reffig{stack_classification}.

The stacking strategy relies on the strength of each individual estimator
using its output as input to a final estimator. The outputs of the classifiers form instances that will be used as input data for a meta-classifier that generates the final classification result. 

Any multi-label learner can be used as a meta-classifier for the Stack method, but in this work the chosen method was the Classifier Chains.
From this point on, the use of the stacking fusion method in conjunction with ECC will be called STACKECC.

\tikzset{every picture/.style={line width=0.5pt}} %set default line width to 0.75pt       

\begin{figure}[H]
    \centering

%\tikzset{every picture/.style={line width=0.75pt}} %set default line width to 0.75pt        

\begin{tikzpicture}[x=0.55pt,y=0.45pt,yscale=-1,xscale=1.1]

%Shape: Rectangle [id:dp6489508267311632] 
\draw   (70,5) -- (140,5) -- (140,96.67) -- (70,96.67) -- cycle ;
%Shape: Circle [id:dp034689680969382186] 
\draw   (379.43,172) .. controls (379.43,158.19) and (390.62,147) .. (404.43,147) .. controls (418.24,147) and (429.43,158.19) .. (429.43,172) .. controls (429.43,185.81) and (418.24,197) .. (404.43,197) .. controls (390.62,197) and (379.43,185.81) .. (379.43,172) -- cycle ;
%Shape: Rectangle [id:dp2948240824947528] 
\draw   (263,126) -- (333,126) -- (333,217.67) -- (263,217.67) -- cycle ;
%Shape: Ellipse [id:dp13786131893759035] 
\draw   (23,172) .. controls (23,160.95) and (60.01,152) .. (105.67,152) .. controls (151.32,152) and (188.33,160.95) .. (188.33,172) .. controls (188.33,183.05) and (151.32,192) .. (105.67,192) .. controls (60.01,192) and (23,183.05) .. (23,172) -- cycle ;
%Straight Lines [id:da8800164707497986] 
\draw    (105.33,96.67) -- (105.65,150) ;
\draw [shift={(105.67,152)}, rotate = 269.65] [color={rgb, 255:red, 0; green, 0; blue, 0 }  ][line width=0.75]    (10.93,-3.29) .. controls (6.95,-1.4) and (3.31,-0.3) .. (0,0) .. controls (3.31,0.3) and (6.95,1.4) .. (10.93,3.29)   ;
%Straight Lines [id:da927348284311226] 
\draw    (188.33,171) -- (260.43,171) ;
\draw [shift={(262.43,171)}, rotate = 180] [color={rgb, 255:red, 0; green, 0; blue, 0 }  ][line width=0.75]    (10.93,-3.29) .. controls (6.95,-1.4) and (3.31,-0.3) .. (0,0) .. controls (3.31,0.3) and (6.95,1.4) .. (10.93,3.29)   ;
%Straight Lines [id:da6672276322152497] 
\draw    (140.2,46.8) -- (261.03,169.57) ;
\draw [shift={(262.43,171)}, rotate = 225.46] [color={rgb, 255:red, 0; green, 0; blue, 0 }  ][line width=0.75]    (10.93,-3.29) .. controls (6.95,-1.4) and (3.31,-0.3) .. (0,0) .. controls (3.31,0.3) and (6.95,1.4) .. (10.93,3.29)   ;
%Straight Lines [id:da04525726262318441] 
\draw    (333.43,172) -- (377.43,172) ;
\draw [shift={(379.43,172)}, rotate = 180] [color={rgb, 255:red, 0; green, 0; blue, 0 }  ][line width=0.75]    (10.93,-3.29) .. controls (6.95,-1.4) and (3.31,-0.3) .. (0,0) .. controls (3.31,0.3) and (6.95,1.4) .. (10.93,3.29)   ;
%Straight Lines [id:da17410230125452086] 
\draw    (429.43,172) -- (459.43,172) ;
\draw [shift={(461.43,172)}, rotate = 180] [color={rgb, 255:red, 0; green, 0; blue, 0 }  ][line width=0.75]    (10.93,-3.29) .. controls (6.95,-1.4) and (3.31,-0.3) .. (0,0) .. controls (3.31,0.3) and (6.95,1.4) .. (10.93,3.29)   ;

% Text Node
\draw (77,6) node [anchor=north west][inner sep=0.75pt]    {$ \begin{array}{l}
(\xx_{1} ,\yy_{1})\\
(\xx_{2} ,\yy_{2})\\
\ \ \ \dotsc \\
(\xx_{n} ,\yy_{n})
\end{array}$};
% Text Node
\draw (270,127) node [anchor=north west][inner sep=0.75pt]    {$ \begin{array}{l}
(\xx '_{1} ,\yy_{1})\\
(\xx '_{2} ,\yy_{2})\\
\ \ \ \dotsc \\
(\xx '_{n} ,\yy_{n})
\end{array}$};
% Text Node
\draw (50,5) node [anchor=north west][inner sep=0.75pt]    {$D$};
% Text Node
\draw (240,125) node [anchor=north west][inner sep=0.75pt]    {$D'$};
% Text Node
\draw (3,138) node [anchor=north west][inner sep=0.75pt]    {$ECC$};
% Text Node
\draw (42,163) node [anchor=north west][inner sep=0.75pt]    {$CC_{1} ,CC_{2} ,\dotsc ,\ CC_{c}$};
% Text Node
\draw (390,160.86) node [anchor=north west][inner sep=0.75pt]    {$CC$};
% Text Node
\draw (362,127.86) node [anchor=north west][inner sep=0.75pt]    {$MC$};

\end{tikzpicture}

    \caption{The Stacking Classifier Ensemble training diagram.}
    \label{fig:stackecc_dg}
\end{figure}
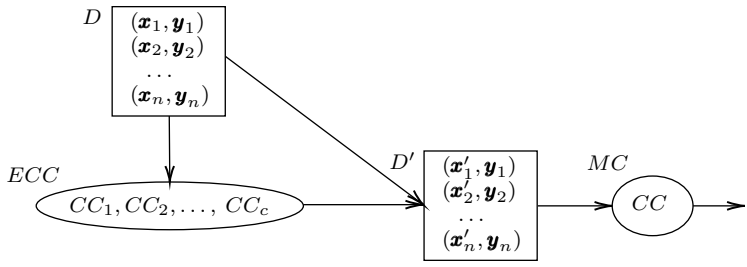

\begin{algorithm}[h]
\newcommand{\forcond}{}
\SetKwFunction{Range}{}
\caption{Stacking Training. Adapted from \cite{tang2014data}. }\label{algorithm:stack}
\SetKwProg{Fn}{Function}{}{end}
\Fn{$\yy \leftarrow$ Stack($D = \{\xx_i,\yy_i\}_{i=1}^{n}$)}
{
	\KwIn{Training data $D$,
	      trained set of classifiers $\hh$}
	\KwOut{An ensemble classifier $\mathcal{E}$}
	
	$c \leftarrow |\hh|$;
	
	Step 1: Learn first level classifiers\\
	
    \For{$i = 1$ to $c$} {
        Learn a base classifier $\hat{h}_i$ based on $D$;
    }
	Step 2: Construct new data sets from $D$
	
    \For{$i = 1$ to $n$} {
        Construct a new data set that contains $\{\xx'_i,\yy_i\}$, where $\xx'_i = \{\hat{h}_1(\xx_i),\hat{h}_2(\xx_i),\hdots,\hat{h}_{c}(\xx_i)\}$;
    }
    Step 3: Learn a second-level classifier \\
    Learn a new classifier $h'$ based on the newly constructed data set
	
    \Return $\mathcal{E}(\xx) = h'(\hat{h}_1(\xx),\hat{h}_2(\xx),\hdots,\hat{h}_{c}(\xx))$
}
\end{algorithm}

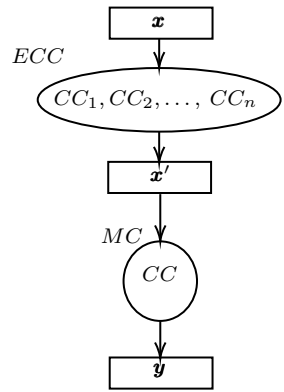
\begin{figure}[h]
    \centering

    \tikzset{every picture/.style={line width=0.75pt}} %set default line width to 0.75pt        
    
    \begin{tikzpicture}[x=0.55pt,y=0.6pt,yscale=-1,xscale=1]
    
    %Shape: Rectangle [id:dp2525225304762855] 
    \draw   (69,18) -- (139,18) -- (139,37.57) -- (69,37.57) -- cycle ;
    %Shape: Rectangle [id:dp8482035028361308] 
    \draw   (68,115) -- (138,115) -- (138,134.57) -- (68,134.57) -- cycle ;
    %Shape: Ellipse [id:dp7560789894841116] 
    \draw   (20,76) .. controls (20,64.95) and (57.01,56) .. (102.67,56) .. controls (148.32,56) and (185.33,64.95) .. (185.33,76) .. controls (185.33,87.05) and (148.32,96) .. (102.67,96) .. controls (57.01,96) and (20,87.05) .. (20,76) -- cycle ;
    %Straight Lines [id:da9962284530585985] 
    \draw    (102.67,96) -- (102.67,113.6) ;
    \draw [shift={(102.67,115.6)}, rotate = 270] [color={rgb, 255:red, 0; green, 0; blue, 0 }  ][line width=0.75]    (10.93,-3.29) .. controls (6.95,-1.4) and (3.31,-0.3) .. (0,0) .. controls (3.31,0.3) and (6.95,1.4) .. (10.93,3.29)   ;
    %Straight Lines [id:da3140675166019975] 
    \draw    (102.67,37.6) -- (102.67,54) ;
    \draw [shift={(102.67,56)}, rotate = 270] [color={rgb, 255:red, 0; green, 0; blue, 0 }  ][line width=0.75]    (10.93,-3.29) .. controls (6.95,-1.4) and (3.31,-0.3) .. (0,0) .. controls (3.31,0.3) and (6.95,1.4) .. (10.93,3.29)   ;
    %Shape: Circle [id:dp6421478819994157] 
    \draw   (78.43,190) .. controls (78.43,176.19) and (89.62,165) .. (103.43,165) .. controls (117.24,165) and (128.43,176.19) .. (128.43,190) .. controls (128.43,203.81) and (117.24,215) .. (103.43,215) .. controls (89.62,215) and (78.43,203.81) .. (78.43,190) -- cycle ;
    %Straight Lines [id:da392660843201446] 
    \draw    (103.43,135.29) -- (103.43,163) ;
    \draw [shift={(103.43,165)}, rotate = 270] [color={rgb, 255:red, 0; green, 0; blue, 0 }  ][line width=0.75]    (10.93,-3.29) .. controls (6.95,-1.4) and (3.31,-0.3) .. (0,0) .. controls (3.31,0.3) and (6.95,1.4) .. (10.93,3.29)   ;
    %Straight Lines [id:da8575808839061569] 
    \draw    (103.43,215) -- (103.43,236.6) ;
    \draw [shift={(103.43,238.6)}, rotate = 270] [color={rgb, 255:red, 0; green, 0; blue, 0 }  ][line width=0.75]    (10.93,-3.29) .. controls (6.95,-1.4) and (3.31,-0.3) .. (0,0) .. controls (3.31,0.3) and (6.95,1.4) .. (10.93,3.29)   ;
    %Shape: Rectangle [id:dp27782097990766896] 
    \draw   (69,238) -- (139,238) -- (139,257.57) -- (69,257.57) -- cycle ;
    
    % Text Node
    \draw (96,22) node [anchor=north west][inner sep=0.75pt]    {$\xx$};
    % Text Node
    \draw (95,115) node [anchor=north west][inner sep=0.75pt]    {$\xx'$};
    % Text Node
    \draw (0,42) node [anchor=north west][inner sep=0.75pt]    {$ECC$};
    % Text Node
    \draw (29,68) node [anchor=north west][inner sep=0.75pt]    {$CC_{1} ,CC_{2} ,\dotsc ,\ CC_{n}$};
    % Text Node
    \draw (89,178.86) node [anchor=north west][inner sep=0.75pt]    {$CC$};
    % Text Node
    \draw (61,155) node [anchor=north west][inner sep=0.75pt]    {$MC$};
    % Text Node
    \draw (97,240) node [anchor=north west][inner sep=0.75pt]    {$\yy$};

    \end{tikzpicture}
    
    \caption{Schematic representation of the Stacking Classifier Ensemble's classification process for an unseen instance.}
    \label{fig:stack_classification}
\end{figure}

\begin{algorithm}[h]
\newcommand{\forcond}{}
\SetKwFunction{Range}{}
\caption{Stacking Classification. }\label{algorithm:classifystack}
\SetKwProg{Fn}{Function}{}{end}
\Fn{$\yy \leftarrow$ ClassifyStack($\xx$,$\hh$)}
{
	\KwIn{feature vector from an unseen instance $\xx$,
	      trained set of classifiers $\hh$}
	\KwOut{Label set $\yy$ assigned to the instance $\xx$ }
	
	$c \leftarrow |\hh|$;
	
	Step 1: Construct new feature vector $\xx'$, where\\
	$\xx' = \{\hat{h}_1(\xx_i),\hat{h}_2(\xx_i),\hdots,\hat{h}_{c}(\xx_i)\}$;\\
	
	Step 2: Classify new feature vector using the ensemble meta-classifier\\
	$\yy \leftarrow \mathcal{E}(\xx'))$\\
	
    \Return $\yy$
}
\end{algorithm}

\subsection{DTECC}
\label{sec:DTECC}

In the DTECC method \cite{freitas2022ensemble}, given an ensemble composed by $c$ multi-label classifiers, the problem is transformed into $m$ binary classification problems. The DTECC method condenses the information of the original $c\times m$ matrix into $c\times1$ DP matrices. The DP for each label $\ell_j \in L,~j \in [1,m]$ is defined as:
\begin{equation}
\text{DP}_{j}(\xx)=\left(
\begin{array}{cc}
d_{1,\ell_{j}}(\xx)  \\
\ldots \\
d_{c,\ell_{j}}(\xx)  \\
%\vdots & \vdots & \ddots & \vdots  \\
\end{array}
\right),~j=1,\ldots,m.
\label{eq:DTECC}
\end{equation}
where $d_{i,\ell_{j}}(\xx)$ represents the confidence level  that the feature vector $\xx$ should be associated with the label $\ell_j$ generated by the i-th classifier. 

Using the previously labeled multi-label training set  $D = \left\{ ({\xx}_{1},\yy_{1}), \ldots, ({\xx}_{n},{\yy}_{n}) \right\}$, 
two corresponding matrices are computed for each label $\ell_j \in L, j=1,\hdots,m$:

\begin{equation}
\text{DT}_j = \frac{ \sum_{k=1}^{n} f(\yy_k,\ell_j) \text{DP}_{j}(\xx_k) }{ \sum_{k=1}^{n} f(\yy_k,\ell_j)},
\label{eq:MDT1}
\end{equation}

\begin{equation}
\text{DT}_{\overline{j}} = \frac{   \left | \sum_{k=1}^{n} (1-f(\yy_k,\ell_j))  \text{DP}_{j}(\xx_k) \right | }{ \sum_{k=1}^{n} \left | 1-f(\yy_k,\ell_j)\right |},
\label{eq:MDT0}
\end{equation}
where $f(\yy_k,\ell_j)$ is a function that takes value 1 if $\ell_j \in \yy_k$. Otherwise, its value is 0. 

$\text{DT}_j$ represents the average DP matrices of the training instances associated with the label $\ell_j$, whilst $\text{DT}_{\overline{j}}$ represents the average DP of the complementary instances. 

The similarity measures $\mu_{j}(\xx)$ and $\mu_{\overline{j}}(\xx)$ are introduced to quantify the alignment between an instance's decision template $\text{DP}(\xx)$ and the reference patterns $\text{DT}_j$ and $\text{DT}_{\overline{j}}$, respectively. Using \refeq{similar} the level of similarity measured between $\text{DP}(\xx)$ and both $\text{DT}_j$ and $\text{DT}_{\overline{j}}$ is computed. Finally, the classification decision rule assigns label $\ell_j$ to instance $\mathbf{x}$ when $\mu_{j}(\xx) > \mu_{\overline{j}}(\xx)$. 

\begin{align}
\mu_{j}(\xx) = 1 - ||\text{DP}_{j}(\xx)-\text{DT}_{j}||_{2}^{2}= \nonumber \\
1 - \sum_{k=1}^{c}
	\left[ d_{k,\ell_j}(\xx) - \text{DT}_{j}(k) \right]^{2}.
\label{eq:similar}
\end{align}

Based on this approach, this paper proposes an alternative version of the DTECC method called UDDTECC. 
 
\section{UDDTECC: Unconditionally Dependent Decision Templates for Ensemble of Classifier Chains}
\label{sec:UDDTECC}

This section  defines the concepts required for understanding the UDDTECC method and suggests a procedure for using it. 

\subsection{UDDTECC Definition}

The proposed UDDTECC method aims to explore correlations between labels by including  additional information regarding labels related to the label that is currently being classified. In addition to the information already used by the DTECC method, the classification process for a label $l_j$ includes in its matrices $DP_j$ and $DT_j$ all labels correlated to $l_j$. Therefore the number of labels included in the DP matrices for each label depends on the number of correlations found with the other labels in the problem.

In order to identify unconditionally related label pairs, Phi ($\phi$) \cite{cohen2013applied} coefficients are used. \
Given two labels $\ell_a $ and $ \ell_b \in L$, and a contingency table for both labels as in \refTab{cont_dep}, $\phi$ coefficient is calculated as: 
\begin{table}[h]
    \centering
\begin{tabular}{r|cc}
           & $\ell_b$ & $\neg\ell_b$ \\ \hline
$\ell_a$     & A      & B           \\
$\neg\ell_a$ & C      & D                                          
\end{tabular}
    \caption{Contingency table for labels $\ell_a$ and $\ell_b$}
    \label{tab:cont_dep}
\end{table}

\begin{equation}
    \phi(\ell_a,\ell_b) = \frac{AD-BC}{\sqrt{(A+B)(C+D)(A+C)(B+D)}}
\end{equation}

Similarly, the Chi ($\chi^2$) coefficient can also be used to determine unconditional relationships between pairs of labels. If $\chi^2 > 6.635 $ values for a label pair, they are considered dependent at 99\% confidence \cite{greenwood1996guide}.

\begin{equation}
    \chi^{2}(\ell_a,\ell_b) = \frac{(AD-BC)^2(A+B+C+D)}{(A+B)(C+D)(A+C)(B+D)} 
\end{equation}

$\chi$ and $\phi$ coefficients follow the mathematical relation $\chi^2(\ell_a,\ell_b) = n \cdot \sqrt{\phi(\ell_a,\ell_b)}$, where $n$ is the number of instances. One advantage of using the $\phi$ coefficient instead of $\chi$ as the threshold parameter for defining the correlations between the labels lies in the range defined by $\phi$.  
As we are working with a $2 \times 2$ contingency matrix, the $\phi$ coefficient is within the $-1$ and $1$ range while the $\chi$ coefficient can vary indefinitely according to the number of evaluated instances. Using the $\phi$ parameter allows a value to be set regardless of the number of instances being evaluated.

Like the DTECC, in the UDDTECC method, the $\text{DT}_j$ and $\text{DT}_{\overline{j}}$ matrices for each label are calculated using  \refeq{MDT1} and \refeq{MDT0} respectively. 

The difference between the proposed methods lies on the $DP$ matrix of each label used for defining both $\text{DT}_j$ and $\text{DT}_{\overline{j}}$ as well as the $\text{DP}_{j}(\xx)$ matrix of the instance $\xx$ to be classified.
A user-defined threshold value $\phi_t$ defines whether or not pairs of labels are correlated. If no correlation is identified among the labels the matrix becomes the same used in the DTECC method. Otherwise, only the uncorrelated labels are kept in the matrix.

Using the previously labeled multi-label training set $D = \left\{ ({\xx}_{1},\yy_{1}), \ldots, ({\xx}_{n},{\yy}_{n}) \right\}$, and $\pmb{L}_j = \{l_p  |  l_p \in L \land|\phi(\ell_p,\ell_j)| \ge \phi_t\}$ representing the labels unconditionally dependent to $\ell_j$, therefore the DP for each label $\ell_j \in L, j=1,\hdots,m$ becomes:
\begin{equation}
\text{DP}_{j}(\xx)=\left(
\begin{array}{cc}
 d_{1,\ell_{p_1}}(\xx) & \ldots\\
 \ldots & \ldots\\
 d_{c,\ell_{p_1}}(\xx) & \ldots \\
%\vdots & \vdots & \ddots & \vdots  \\
\end{array}
\right), j=1,\ldots,m  \hspace{.25cm} \textit{and}  \hspace{.25cm} \{\ell_{p_1}, \, \ldots\} = \pmb{L}_j
\label{eq:dteccddp}
\end{equation}

\begin{figure}
	\centering
	\includegraphics[width=.6\textwidth]{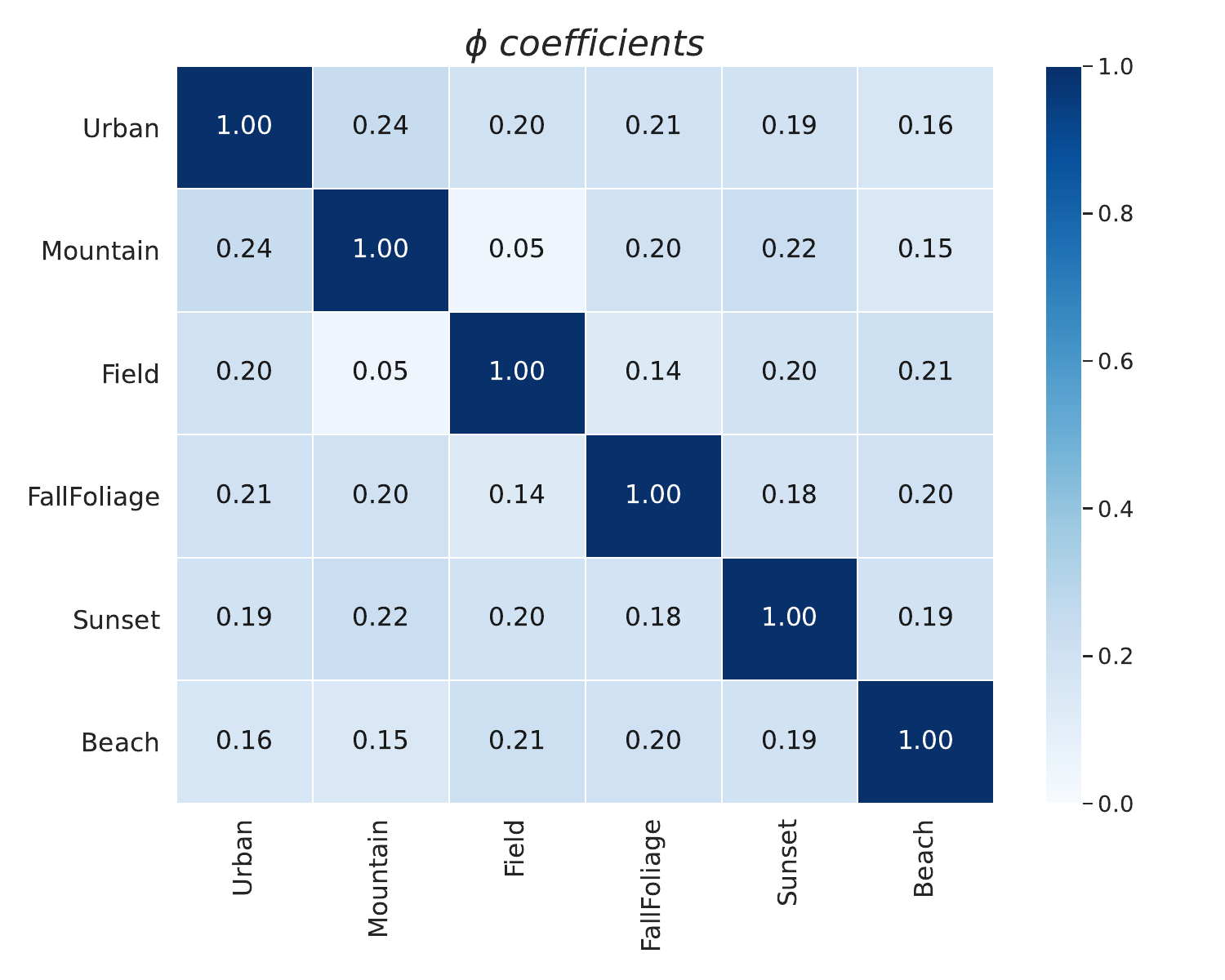} 
	\caption{ 
	Matrix of the $\phi$ coefficients absolute values of the Scene dataset labels.}
	\label{fig:proposta-scenecorr}
\end{figure}

Let's take the $\phi$ coefficients matrix for the Scene dataset shown in \reffig{proposta-scenecorr} as an example. Intuitively, the $\phi$ coefficients matrix is symmetric because each pair of variables has the same relationship whether its $\phi$ coefficient is in the upper right triangle or the lower left triangle. If a value of $\phi_t = 0.20$ is set, the Mountain class would be considered unconditionally dependent to the Sunset and Urban classes in the fusion process. In this case, the vector of unconditionally related labels to Mountain would be $P_{Mountain} = \{Mountain, Sunset, Urban\}$ and the DP matrix for the label Mountain considered would become:

\begin{equation}
\text{DP}_{Mountain}(\xx)=\left(
\begin{array}{ccc}
d_{1,Mountain}(\xx)  & d_{1,Sunset}(\xx) & d_{1,Urban}(\xx)\\
\ldots & \ldots & \ldots\\
d_{c,Mountain}(\xx)  & d_{c,Sunset}(\xx) & d_{c,Urban}(\xx) \\
\end{array}
\right)
\label{eq:dteccddpm}
\end{equation}

Given these conditions, the classification of an unseen instance is exemplified in \reffig{ddtecc_ce}.

\tikzset{every picture/.style={line width=0.75pt}} %set default line width to 0.75pt        

\begin{figure}
    \centering
    \begin{tikzpicture}[x=0.5pt,y=0.5pt,yscale=-1,xscale=1]
%uncomment if require: \path (0,321); %set diagram left start at 0, and has height of 321

%Straight Lines [id:da7439176481247629] 
\draw    (107.33,157) -- (125,157) ;
%Straight Lines [id:da908391548891718] 
\draw    (125,157) -- (125,97) ;
%Straight Lines [id:da6975031926323105] 
\draw    (125,157) -- (125,241) ;
%Straight Lines [id:da38089764807567894] 
\draw    (125,241) -- (167,241) ;
\draw [shift={(169,241)}, rotate = 180] [color={rgb, 255:red, 0; green, 0; blue, 0 }  ][line width=0.75]    (10.93,-3.29) .. controls (6.95,-1.4) and (3.31,-0.3) .. (0,0) .. controls (3.31,0.3) and (6.95,1.4) .. (10.93,3.29)   ;
%Straight Lines [id:da9749044228513595] 
\draw    (125,97) -- (163,97) ;
\draw [shift={(165,97)}, rotate = 180] [color={rgb, 255:red, 0; green, 0; blue, 0 }  ][line width=0.75]    (10.93,-3.29) .. controls (6.95,-1.4) and (3.31,-0.3) .. (0,0) .. controls (3.31,0.3) and (6.95,1.4) .. (10.93,3.29)   ;
%Straight Lines [id:da3582703729043928] 
\draw    (272,97) -- (512,97) ;
%Straight Lines [id:da6165038254447366] 
\draw    (274,242) -- (512,242) ;
%Shape: Ellipse [id:dp07335419546379085] 
\draw   (477,158) .. controls (477,146.95) and (492.67,138) .. (512,138) .. controls (531.33,138) and (547,146.95) .. (547,158) .. controls (547,169.05) and (531.33,178) .. (512,178) .. controls (492.67,178) and (477,169.05) .. (477,158) -- cycle ;
%Straight Lines [id:da467659260663591] 
\draw    (512,97) -- (512,138) ;
%Straight Lines [id:da6737209131808766] 
\draw    (512,178) -- (512,242) ;
%Straight Lines [id:da3573687905995473] 
\draw    (547,158) -- (561,158) ;
\draw [shift={(563,158)}, rotate = 180] [color={rgb, 255:red, 0; green, 0; blue, 0 }  ][line width=0.75]    (10.93,-3.29) .. controls (6.95,-1.4) and (3.31,-0.3) .. (0,0) .. controls (3.31,0.3) and (6.95,1.4) .. (10.93,3.29)   ;
%Shape: Rectangle [id:dp8415597591770452] 
\draw   (40.5,100.5) -- (68.5,100.5) -- (68.5,213.5) -- (40.5,213.5) -- cycle ;
%Shape: Rectangle [id:dp094540393432901] 
\draw   (8.5,100.5) -- (40.5,100.5) -- (40.5,213.5) -- (8.5,213.5) -- cycle ;
%Shape: Rectangle [id:dp40626948346294256] 
\draw   (68.5,100.5) -- (100.5,100.5) -- (100.5,213.5) -- (68.5,213.5) -- cycle ;
%Straight Lines [id:da7966996138277909] 
\draw    (23,213) -- (23,273.5) ;
\draw [shift={(23,275.5)}, rotate = 270] [color={rgb, 255:red, 0; green, 0; blue, 0 }  ][line width=0.75]    (10.93,-3.29) .. controls (6.95,-1.4) and (3.31,-0.3) .. (0,0) .. controls (3.31,0.3) and (6.95,1.4) .. (10.93,3.29)   ;
%Straight Lines [id:da12285601079357455] 
\draw    (86,213) -- (86,232.5) ;
\draw [shift={(86,234.5)}, rotate = 270] [color={rgb, 255:red, 0; green, 0; blue, 0 }  ][line width=0.75]    (10.93,-3.29) .. controls (6.95,-1.4) and (3.31,-0.3) .. (0,0) .. controls (3.31,0.3) and (6.95,1.4) .. (10.93,3.29)   ;
%Straight Lines [id:da8421934096649091] 
\draw    (54,214) -- (54,255.5) ;
\draw [shift={(54,257.5)}, rotate = 270] [color={rgb, 255:red, 0; green, 0; blue, 0 }  ][line width=0.75]    (10.93,-3.29) .. controls (6.95,-1.4) and (3.31,-0.3) .. (0,0) .. controls (3.31,0.3) and (6.95,1.4) .. (10.93,3.29)   ;

% Text Node
\draw (-1,105.4) node [anchor=north west][inner sep=0.75pt]    {$\begin{pmatrix}
0.8 & 0.3 & 0.7\\
0.7 & 0.2 & 0.8\\
0.2 & 0.8 & 0.9\\
0.9 & 0.1 & 0.8\\
0.3 & 0.5 & 0.8
\end{pmatrix}$};
% Text Node
\draw (0,77.4) node [anchor=north west][inner sep=0.75pt]    {$DP_{Mountain}$};
% Text Node
\draw (165,52.4) node [anchor=north west][inner sep=0.75pt]    {$\begin{pmatrix}
0.9 & 0.1 & 0.3\\
0.9 & 0.2 & 0.3\\
0.1 & 0.9 & 0.8\\
0.8 & 0.3 & 0.7\\
0.3 & 0.6 & 0.8
\end{pmatrix}$};
% Text Node
\draw (166,193.4) node [anchor=north west][inner sep=0.75pt]    {$\begin{pmatrix}
0.2 & 0.2 & 0.3\\
0.1 & 0.7 & 0.7\\
0.5 & 0.7 & 0.3\\
0.2 & 0.9 & 0.2\\
0.2 & 0.6 & 0.3
\end{pmatrix}$};
% Text Node
\draw (168,30.4) node [anchor=north west][inner sep=0.75pt]    {$DT_{Mountain}$};
% Text Node
\draw (169,169.4) node [anchor=north west][inner sep=0.75pt]    {$DT_{\overline{Mountain}}$};
% Text Node
\draw (274,72.4) node [anchor=north west][inner sep=0.75pt]    {$\mu _{1} =1-||DP_{Mountain} -DT_{Mountain} ||_{2}^{2} \ =0.4\ $};
% Text Node
\draw (276,245.4) node [anchor=north west][inner sep=0.75pt]    {$\mu _{\overline{1}} =1-||DP_{Mountain} -DT_{\overline{Mountain}} ||_{2}^{2} =-2.37\ $};
% Text Node
\draw (480,146.4) node [anchor=north west][inner sep=0.75pt]    {$\mu _{1}  >\mu _{\overline{1}} \ $};
% Text Node
\draw (563,150.4) node [anchor=north west][inner sep=0.75pt]    {$\yy_{Mountain} =1$};
% Text Node
\draw (44,255.4) node [anchor=north west][inner sep=0.75pt]    {$\ell _{Sunset}$};
% Text Node
\draw (80,235.4) node [anchor=north west][inner sep=0.75pt]    {$\ell _{Urban}$};
% Text Node
\draw (10,274.4) node [anchor=north west][inner sep=0.75pt]    {$\ell _{Mountain}$};

\end{tikzpicture}
    \caption{Example of classifier fusion using the UDDTECC method on an instance of the
Scene data set for the label Mountain.}
    \label{fig:ddtecc_ce}
\end{figure}
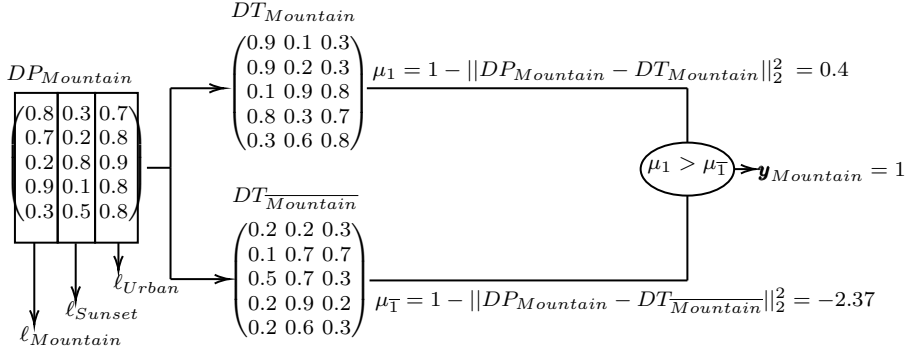

 The similarity value between the $\text{DP}_j$ matrix of the label $\ell_j$ and the $\text{DT}_j$ and $\text{DT}_{\overline{j}}$ matrices are calculated using \refeq{similard}. As in the original DTECC method, an instance is labeled as $\ell_j$ only if $\mu_{j}(\xx) > \mu_{\overline{j}}(\xx)$.

\begin{align}
\mu_{j}(\xx) = 
1 - \sum_{i=1}^{c} \, \sum_{\ell_k \in L_{j}}{}
	\left[ d_{i,\ell_k}(\xx) - \text{DT}_{j}({i,k}) \right]^{2}.
\label{eq:similard}
\end{align}

\subsection{UDDTECC Workflow}

Since the UDDTECC method is a fusion method that requires a complex training process, \reffig{uddtecflow} describes its workflow.
First, the available training data should be divided into an inner training and an inner validation subsets.
Then, the $\phi$ coefficients matrix is computed using the inner training subset. The next step consists of tuning the $\phi_t$ threshold value. For each $\phi_t$ candidate  value, an UDDTECC model is trained using the inner training subset and its performance is evaluated on the inner validation subset. Evaluation is performed by classifying all inner validation data and computing a classification  metric. 
The best performing candidate $\phi_t$ value should be chosen as the $\phi_t$. Once the value of $\phi_t$ is selected, the next step consists of using all available training data (using both inner training and inner validation subsets) with the chosen $\phi_t$. In the last step, the classifier may be applied to unseen instances.

\tikzstyle{process} = [rectangle, minimum width=3cm, minimum height=1cm, text centered, draw=black, fill=orange!30]
\tikzstyle{startstop} = [rectangle, rounded corners, minimum width=3cm, minimum height=1cm,text centered, draw=black, fill=red!30]
\tikzstyle{io} = [trapezium, trapezium left angle=70, trapezium right angle=110, minimum width=3cm, minimum height=1cm, text centered, draw=black, fill=blue!30]
\tikzstyle{decision} = [diamond, minimum width=3cm, minimum height=1cm, text centered, draw=black, fill=green!30]

\tikzstyle{arrow} = [thick,->,>=stealth]

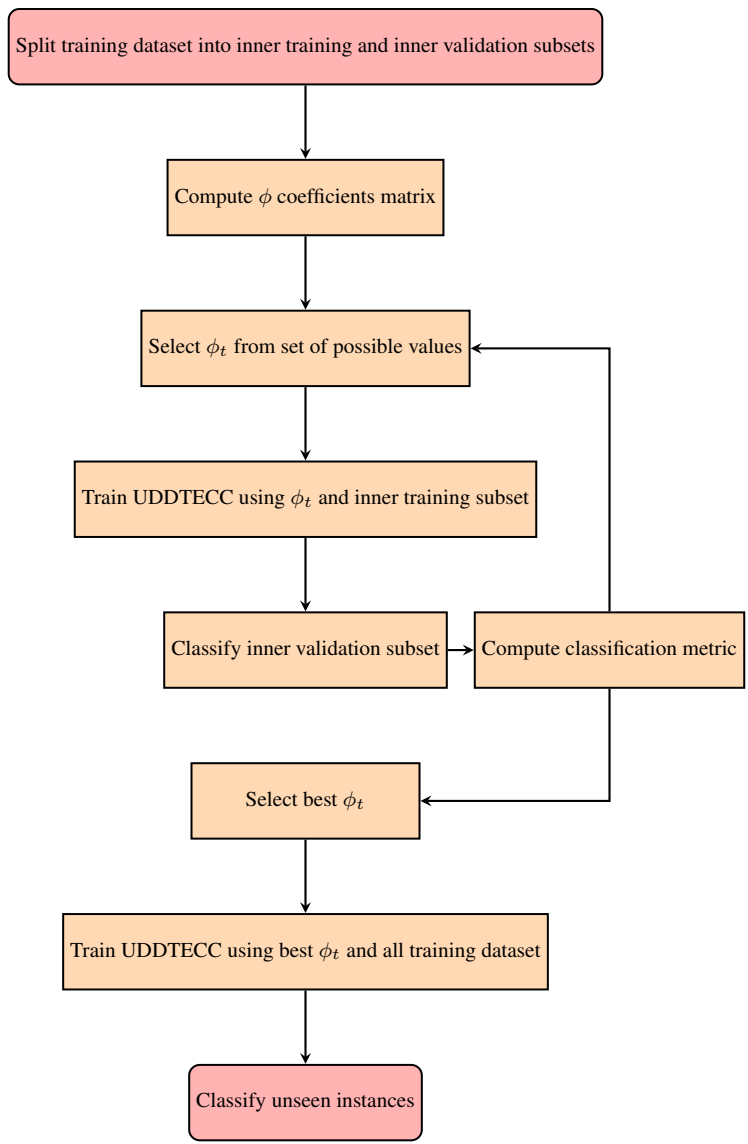
\begin{figure}[t]
    \centering

    \begin{tikzpicture}[node distance=2cm]
    \node (start) [startstop] {Split training dataset into inner training and inner validation subsets};
    \node (in1) [process, below of=start] {Compute $\phi$ coefficients matrix};
    \node (pro1) [process, below of=in1] {Select $\phi_t$ from set of possible values};
    \node (dec1) [process, below of=pro1] {Train UDDTECC using $\phi_t$ and inner training subset};
    \node (dec2) [process, below of=dec1, xshift=0cm] {Classify inner validation subset};
    \node (pro2a) [process, right of=dec2, xshift=2cm] {Compute classification metric};
    \node (pro2b) [process, below of=dec2] {Select best $\phi_t$};
    \node (out1) [process, below of=pro2b] {Train UDDTECC using best $\phi_t$ and all training dataset};
    \node (stop) [startstop, below of=out1] {Classify unseen instances};
    \draw [arrow] (start) -- (in1);
    \draw [arrow] (in1) -- (pro1);
    \draw [arrow] (pro1) -- (dec1);
    \draw [arrow] (dec1) -- (dec2);
    \draw [arrow] (dec2) -- (pro2a);
    \draw [arrow] (pro2a) |- (pro2b);
    \draw [arrow] (pro2a) |- (pro1);
    \draw [arrow] (pro2b) -- (out1);
    \draw [arrow] (out1) -- (stop);
    \end{tikzpicture}

    \caption{UDDTECC training workflow.}
    \label{fig:uddtecflow}
\end{figure}

\section{Experiments} \label{sec:setup}

The experiments evaluated UDDTECC fusion method, the original DTECC method, Majority Vote (MVECC), Mean Ensemble (MEECC) and a Stacking ECC (STACKECC) method. All classifiers used in the experiments are evaluated using a 10-fold cross-validation technique. 

Our experimental evaluation employs datasets spanning multiple domains, with detailed characteristics provided in \reftab{datasets}. The diversity of a multi-label dataset, measured as the percentage of observed unique label combinations relative to all possible combinations, serves as a key indicator of label distribution complexity. When the diversity value approaches its maximum (near 1), the dataset contains nearly all possible label combinations, indicating rich and varied co-occurrence patterns that typically present greater learning challenges. Conversely, datasets with low diversity values contain only a limited subset of possible label combinations, resulting in more predictable and homogeneous annotation patterns. The total number of possible combinations grows exponentially with the number of labels, making diversity an essential metric for understanding the complexity of multi-label interactions in different domains. The selection prioritized computationally feasible cases while maintaining diversity in label interactions, excluding larger datasets and complex feature spaces due to hardware limitations in our current Mulan/Weka implementation.

\begin{table}[H]
    \centering
\begin{tabular}{lrrrr}
\toprule
Name                   & Features & Diversity & Instances  & Labels \\ \midrule
3sources\_guardian1000 & 1000                             & 0.219                             & 302                             & 6                            \\
3sources\_inter3000    & 3000                             & 0.172                             & 169                             & 6                            \\ 
CAL500                 & 68                               & 1.000                            & 502                             & 174                          \\ 
Emotions               & 72                               & 0.422                             & 593                             & 6                            \\ 
Enron                  & 1001                             & 0.442                             & 1702                            & 53                           \\ 
Genbase                & 1186                             & 0.048                             & 662                             & 27                           \\ 
GnegativeGO            & 1717                             & 0.074                             & 1392                            & 8                            \\ 
GpositiveGO            & 912                              & 0.438                             & 519                             & 4                            \\ 
Image                  & 294                              & 0.625                             & 2000                            & 5                            \\ 
Langlog                & 1004                             & 0.208                              & 1460                            & 75                           \\ 
Medical                & 1449                             & 0.096                             & 978                             & 45                           \\  
Scene                  & 294                              & 0.234                             & 2407                            & 6                            \\ 
VirusGO                & 749                              & 0.266                             & 207                             & 6                            \\ 
Water-quality          & 16                               & 0.778                             & 1060                            & 14                           \\ 
Yeast                  & 103                              & 0.082                             & 2417                            & 14                           \\ \bottomrule
\end{tabular}
\caption{Characteristics of the multi-label benchmark datasets: showing the number of features, label diversity, instances, and  labels)}
\label{tab:datasets}
\end{table}

Mulan \cite{tsoumakas2011mulan} provides implementations for the Ensemble of Classifier Chains and, for all metrics evaluated, these implementations were used to perform the experiments. 
Reproducibility and verifiability are very important issues when it comes to experimental results \cite{rauber2020experimental}. So, in order to make the experiments in this work reproducible, the code and the results of the experiments are made available in the public repository \href{https://github.com/vfrocha/ddtecc}{https://github.com/vfrocha/ddtecc}.

\subsection{Experimental Methodology}

To ensure an unbiased evaluation of the methods, all experiments use $n = 50$ classifiers in the ensemble and Naive Bayes as a single-label
base classifier. Bayesian classifiers are often efficient when used as base classifiers of ensembles \cite{antonucci2013ensemble}. The STACKECC method uses Classifier Chains as its meta-classifer. Additional ECC parameters were defined as the default values defined by their authors, e.g., size of each bag sample as a percentage of the training size is set to 100, and the choice of sampling with replacement. 
By keeping the values of these parameters constant, one may get a better idea of the impact of the fusion scheme on the final classification result, as this is the only difference that exists between the methods used in the experiments. 

MVECC and MEECC use  a threshold value of $t = 0.5$ as it is usual for these methods. and STACKECC, DTECC and UDDTECC methods do not require this threshold value. 

For the UDDTECC method,  in each iteration of the cross-validation, one of these folds is selected as the test set and the remaining folds are divided into validation and training sets in the ratio of 33\% (3 folds) and 67\% (6 folds) respectively as shown in \reffig{experimantal_folds}. Each UDDTECC model with $\phi_t \in [0.0, 0.25, 0.5, 0.75, 1]$ is then trained on the training set and validated on the validation set. The results for each model are then compared according to a pre-selected metric that is currently being evaluated (Accuracy, F-Measure, Subset Accuracy, and Hamming Loss). The best model is then selected and trained using both the inner training and the inner validation folds and finally applied to the test fold. After 10 iterations we have the results for all folds in the original data set for the selected metric. The final result of the experiment is the average of the metric over these 10 results. 

The UDDTECC performance evaluation framework is shown in \refalg{frame}. The total amount of $n$ patterns is randomly split into $K$ stratified folds. 
Before the training of the classifier is performed,
the best $\phi_t$ hyperparameter for the UDDTECC classifier is estimated in a tuning step.
Three of the training folds are used as inner validation and the remaining folds are used as inner training.
Each evaluated $\phi_t$ value is trained on the inner training folds and validated on the inner validation fold. 
The $\phi_t$ value of the classifier with the best result on the evaluated metric is then selected and used in the cross-validation loop.

\tikzset{every picture/.style={line width=0.75pt}} %set default line width to 0.75pt        
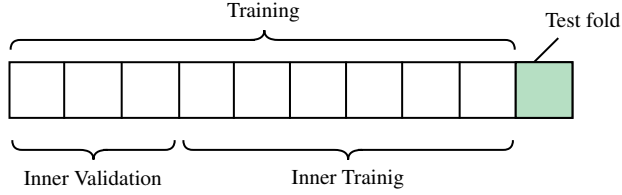
\begin{figure}[h]
    \centering
\begin{tikzpicture}[x=0.75pt,y=0.75pt,yscale=-.7,xscale=.7]

%Shape: Rectangle [id:dp32449231953526025] 
\draw   (77,53) -- (478.43,53) -- (478.43,93) -- (77,93) -- cycle ;
%Straight Lines [id:da08426966981225936] 
\draw    (116,52.71) -- (116,93.71) ;
%Straight Lines [id:da9503922387382591] 
\draw    (157,52.71) -- (157,93.71) ;
%Straight Lines [id:da4974561133807345] 
\draw    (198,52.71) -- (198,93.71) ;
%Straight Lines [id:da6273723801599334] 
\draw    (237,52.71) -- (237,93.71) ;
%Straight Lines [id:da3285536513717826] 
\draw    (277,52.71) -- (277,93.71) ;
%Straight Lines [id:da6834191288787215] 
\draw    (317,52.5) -- (317,93.5) ;
%Straight Lines [id:da07882282731929746] 
\draw    (357,52.5) -- (357,93.5) ;
%Straight Lines [id:da6226287659310108] 
\draw    (398,52.5) -- (398,93.5) ;
%Straight Lines [id:da0046767398734759436] 
\draw    (438,52.5) -- (438,93.5) ;
%Shape: Brace [id:dp8551912155490398] 
\draw   (437.29,44) .. controls (437.29,39.33) and (434.96,37) .. (430.29,37) -- (268.75,37) .. controls (262.08,37) and (258.75,34.67) .. (258.75,30) .. controls (258.75,34.67) and (255.42,37) .. (248.75,37)(251.75,37) -- (84.8,37) .. controls (80.13,37) and (77.8,39.33) .. (77.8,44) ;
%Straight Lines [id:da8092248313230745] 
\draw    (451,53) -- (468.33,35.67) ;
%Shape: Brace [id:dp1305110291392837] 
\draw   (201,109) .. controls (201,113.67) and (203.33,116) .. (208,116) -- (308.5,116) .. controls (315.17,116) and (318.5,118.33) .. (318.5,123) .. controls (318.5,118.33) and (321.83,116) .. (328.5,116)(325.5,116) -- (429,116) .. controls (433.67,116) and (436,113.67) .. (436,109) ;
%Shape: Brace [id:dp4523401310130255] 
\draw   (79,109) .. controls (79,113.67) and (81.33,116) .. (86,116) -- (126.9,116) .. controls (133.57,116) and (136.9,118.33) .. (136.9,123) .. controls (136.9,118.33) and (140.23,116) .. (146.9,116)(143.9,116) -- (187.8,116) .. controls (192.47,116) and (194.8,113.67) .. (194.8,109) ;
%Shape: Rectangle [id:dp07774074987851387] 
\draw  [fill={rgb, 255:red, 182; green, 223; blue, 195 }  ,fill opacity=1 ] (438,53) -- (478.43,53) -- (478.43,93) -- (438,93) -- cycle ;

% Text Node
\draw (230,8) node [anchor=north west][inner sep=0.75pt]   [align=left] {Training};
% Text Node
\draw (457,15) node [anchor=north west][inner sep=0.75pt]   [align=left] {Test fold};
% Text Node
\draw (276,128) node [anchor=north west][inner sep=0.75pt]   [align=left] {Inner Trainig};
% Text Node
\draw (85,129) node [anchor=north west][inner sep=0.75pt]   [align=left] {Inner Validation};

\end{tikzpicture}
    \caption{Fold division used in the experiments.}
    \label{fig:experimantal_folds}
\end{figure}

%--------------------
\begin{algorithm}%[!htb]
\newcommand{\forcond}{}
\SetKwFunction{Range}{}
\caption{The UDDTECC Performance Evaluation Framework. }\label{algorithm:frame}
\SetAlgoLined
\SetKwProg{Fn}{Function}{}{end}
\Fn{PM $\leftarrow$ ModelPerformance($\mathcal{D}$,K)}
{
	\KwIn{$n$ labelled patterns $\mathcal{D}$ from $c$ classes, number of folds $K$ (outer loop) }
	\KwOut{Make $K$ stratified folds.
		One fold is fixed as test, remaining folds are train and tune.
		Calculate performance criterion for each combination of round and fold and put
		into
		$R\times K$ performance matrix PM}

	// generate $K$ stratified folds\\
	$f_{k};  k=1,\ldots,K$\;
	\For( // all folds){$k=1$ \KwTo K} {
		// training set of $k$-th fold \\
		$\mathcal{T}\leftarrow \left\{ f_{1} \cup \ldots \cup f_{K}\right\}
		\setminus f_{k}$ \;
		%$\mathcal{T}_{r}^{\text{TR}}\leftarrow \left\{ f_{r,1} \cup \ldots \cup f_{r,K}
		%\right\} \setminus f_{r,k}$ \;

		// test set of $k$-th fold \\
		$\mathcal{V}\leftarrow f_{k}$ \;

		$\mathcal{P}^{*}\leftarrow$ Tuning($\mathcal{T}$)\;

		%$\mathcal{P}^{*}\leftarrow$ Tuning($\mathcal{T}_{r}^{\text{TR}}$,K-1)\;
		%\textcolor{red}{colocar retorno do classificador e passa como parametro abaixo.}
		$\mathcal{C} \leftarrow $ trainclassifier( $\mathcal{T}, \mathcal{P}^{*}$ )\;
		$crit \leftarrow$ testclassifier($\mathcal{C}, \mathcal{V}$)\;
		PM(k)$\leftarrow crit$
	}
}
\hrulefill\\
\Fn{$\mathcal{P}^{*}\leftarrow$ Tuning($\mathcal{D}$)}
{
	\KwIn{Data set $\mathcal{D}$}
	\KwOut{Optimal hyperparameter set $\mathcal{P}^{*}$ of classifier model}

initialize best criterion $maxcrit\leftarrow 0$\;

// three first folds are used as inner validation\\
$\mathcal{V}\leftarrow$ $\mathcal{D}$[:3]\;
// remainder are used as inner training\\
$\mathcal{T}\leftarrow$ $\mathcal{D}$[3:]\;

\Repeat( // grid search for best hyperparameter set $\mathcal{P}^{*}$){search completed} {
	generate hyperparameter candidate set $\mathcal{P}$\;

		$\mathcal{C} \leftarrow $ trainclassifier( $\mathcal{T}, \mathcal{P}$ )\;
		$crit \leftarrow$ testclassifier($\mathcal{C}, \mathcal{V}$)\;

	%$\overline{crit} \leftarrow mean(crit_{k})$\;
	\If{$crit > maxcrit$}{$maxcrit\leftarrow crit$;\\
		$\mathcal{P}^{\ast}\leftarrow\mathcal{P}$}
}
}
\end{algorithm}

\subsection{Evaluation Metrics}

The evaluation of multi-label methods is performed using six example-based multi-label metrics. Example-based metrics perform the calculation individually for each example and the final result is the average for all examples \cite{zhang2013review}.

The ratio of correctly classified labels to the total number of real and predicted labels in the test sample is called example-based accuracy.

\begin{equation}
	\text{Accuracy}(\pmb{y},\pmb{h}(\pmb{x}))=\frac{ | \pmb{y} 	\cap \pmb{h}(\pmb{x}) | }{ | \pmb{y} 	\cup \pmb{h}(\pmb{x}) | }.
\label{eq:accuracy}
\end{equation}

The fraction of incorrectly classified labels over the total number of labels in the problem is called Hamming Loss \cite{schapire2000boostexter}. Unlike the other metrics used in this paper, Hamming Loss is a loss function and, therefore, should be minimized by the classifiers.

\begin{equation}
	\text{HammingLoss}(\pmb{y},\pmb{\hat{h}}(\pmb{x}))=	\frac{1}{m}\sum _{i=1}^m \left[y_i\neq \hat{h}_i(\pmb{x})\right]
\label{eq:hammingLoss}
\end{equation}

The Subset accuracy, or $\text{Subset}_{0/1}$ accuracy, metric \cite{cheng2010bayes}  considers as correct only those cases where the label set predicted by the classifier $\pmb{h}$ exactly matches the actual label set of the example $\yy$.  All partially correct answers are considered equally wrong. Because it equally penalizes totally wrong and partially correct answers, it is considered a very strict metric.

\begin{equation}
    \text{Subset}_{0/1}(\pmb{y},\pmb{h}(\pmb{x}))=[\pmb{y} = \pmb{h}(\pmb{x})]
\label{eq:subset01}
\end{equation}

Example-Based Recall is the ratio of labels present in $\yy$ that were predicted by $\pmb{h}$.

\begin{equation} \label{eq:recall}
Recall(\pmb{y},\pmb{h}(\pmb{x}))=\frac{ | \pmb{y} \cap \pmb{h}(\pmb{x}) | }{ | \pmb{y} | }.
\end{equation}

Example-Based Precision is the ratio of labels correctly predicted by $\pmb{h}$, i.e., the ratio of labels predicted by $\pmb{h}$ which are present in $\yy$.

\begin{equation}
Precision(\pmb{y},\pmb{h}(\pmb{x}))=\frac{ | \pmb{y} 	\cap \pmb{h}(\pmb{x}) | }{ | \pmb{h}(\pmb{x}) | }.
  \label{eq:precision}
\end{equation}

The Example Based $\text{F}$-measure, also called $\text{F}_1$, is the harmonic mean of Example-Based Precision and Example-Based Recall. In general, a good classifier should get good results in both Example Based Precision and Example Based Recall. The Example-Based F Measure provides a comprehensive evaluation of Precision and Recall. It is  widely used in problems involving unbalanced datasets.

\begin{equation}
F_1(\pmb{y},\pmb{h}(\pmb{x}))=2\cdot \frac{Precision \cdot Recall}{ Precision + Recall }.
  \label{eq:F1}
\end{equation}

\subsection{Overall Results}
\label{sec:odr}

Tables \ref{tab:accuracy}, \ref{tab:fmeasure}, \ref{tab:subsetaccuracy} and \ref{tab:hammingloss} show the results obtained by each method using Accuracy, F-measure, Subset Accuracy and Hamming Loss metrics, respectively. Altogether with the average performance, the rank achieved by the method is also shown in parenthesis. The best results are highlighted in bold face. The average ranking is presented in \refTab{avg_all}. At first glance, one may see that there is no method that is the best in all data sets when evaluated by one of the metrics. %Indeed, every method is the best for at least two data sets in each metric.

\begin{table}[H]
    \centering
\begin{tabular}{llllll}
\toprule
{} &        DTECC &        MEECC &        MVECC &     STACKECC &      UDDTECC \\
Dataset          &              &              &              &              &              \\
\midrule
3sources\_bbc1000      &  0.1497(4.5) &  0.1526(3.0) &  0.1547(2.0) &  \bf{0.2852(1.0)} &  0.1497(4.5) \\
3sources\_guardian1000 &  0.0778(3.5) &  0.0790(2.0) &  0.0773(5.0) & \bf{ 0.2243(1.0)} &  0.0778(3.5) \\
3sources\_inter3000    &  0.0059(4.0) &  0.0088(2.0) &  0.0059(4.0) & \bf{0.1538(1.0)} &  0.0059(4.0) \\
CAL500                &  0.2186(3.0) &  0.2210(2.0) &  \bf{0.2212(1.0)} &  0.1991(5.0) &  0.2175(4.0) \\
Emotions              &  \bf{0.5329(1.0)} &  0.5322(2.0) &  0.5301(4.0) &  0.5141(5.0) &  0.5307(3.0) \\
Enron                 &  0.2484(2.0) &  0.2448(3.0) &  0.2444(4.0) &  0.0863(5.0) &  \bf{0.2503(1.0)} \\
Genbase               &  0.0982(5.0) &  0.2980(3.0) &  0.3007(2.0) &  \bf{0.3819(1.0)} &  0.0994(4.0) \\
GnegativeGO           &  \bf{0.8990(2.0)} &  0.8983(4.0) &  \bf{0.8990(2.0)} &  0.5649(5.0) &  \bf{0.8990(2.0)} \\
GpositiveGO           &  0.8950(4.0) &  0.8931(5.0) &  0.8959(3.0) &  0.9085(2.0) &  \bf{0.9113(1.0)} \\
Langlog               & 0.1958(4.0) & 0.2336(3.0) & 0.2338(2.0) & 0.0513(5.0) & \bf{0.2343(1.0)} \\
Image                 &  0.3579(3.0) &  0.3571(5.0) &  0.3573(4.0) &  0.3614(2.0) &  \bf{0.3652(1.0)} \\
Medical               &  0.0253(5.0) &  0.3714(2.0) &  \bf{0.3736(1.0)} &  0.0458(3.0) &  0.0340(4.0) \\
Scene                 &  \bf{0.4617(1.5)} &  0.4600(3.0) &  0.4589(4.0) &  0.4074(5.0) &  \bf{0.4617(1.5)} \\
VirusGO               &  0.8121(2.5) &  0.8008(4.0) &  \bf{0.8123(1.0)} &  0.6963(5.0) &  0.8121(2.5) \\
Water-quality         &  \bf{0.3923(1.5)} &  0.3870(3.0) &  0.3869(4.0) &  0.3760(5.0) &  \bf{0.3923(1.5)} \\
Yeast                 &  0.4253(4.0) &  0.4294(2.0) &  \bf{0.4309(1.0)} &  0.3821(5.0) &  0.4254(3.0) \\
\bottomrule
\end{tabular}
    \caption{Results for Accuracy for the ECC using all the fusion schemes and data sets.}
    \label{tab:accuracy}
\end{table}

One may also note from \refTab{accuracy} that the UDDTECC method obtained the best Accuracy result in 6 of the 15 datasets evaluated. In second place was the MVECC method with 5 wins, which, together with consistent results in the other datasets, gave it the second best average ranking for Accuracy. DTECC was the best in 4 datasets. The MEECC method did not obtain the best result in any of the evaluated datasets, but its average results in most datasets gave it a better average ranking than the STACKECC method, which, despite obtaining the best result in 4 of the evaluated datasets, was the worst overall. 

\begin{table}[H]
    \centering
\begin{tabular}{llllll}
\toprule
{} &        DTECC &        MEECC &        MVECC &     STACKECC &      UDDTECC \\
Dataset          &              &              &              &              &              \\
\midrule
3sources\_bbc1000      &  0.1640(4.5) &  0.1669(3.0) &  0.1699(2.0) &  \bf{0.3618(1.0)} &  0.1640(4.5) \\
3sources\_guardian1000 &  0.0859(3.5) &  0.0866(2.0) &  0.0854(5.0) &  \bf{0.3053(1.0)} &  0.0859(3.5) \\
3sources\_inter3000    &  0.0078(4.0) &  0.0098(2.0) &  0.0078(4.0) &  \bf{0.2238(1.0)} &  0.0078(4.0) \\
CAL500                &  0.3447(3.0) &  0.3477(2.0) &  \bf{0.3483(1.0)} &  0.3217(5.0) &  0.3428(4.0) \\
Emotions              &  \bf{0.6295(1.0)} &  0.6278(3.0) &  0.6270(4.0) &  0.6175(5.0) &  0.6279(2.0) \\
Enron                 &  0.3649(2.0) &  0.3601(3.0) &  0.3596(4.0) &  0.1554(5.0) &  \bf{0.3659(1.0)} \\
Genbase               &  0.1594(5.0) &  0.3071(3.0) &  0.3104(2.0) &  \bf{0.4339(1.0)} &  0.1605(4.0) \\
GnegativeGO           &  \bf{0.9173(2.0)} &  0.9164(4.0) &  0.9173(2.0) &  0.6841(5.0) &  \bf{0.9173(2.0)} \\
GpositiveGO           &  0.9043(4.0) &  0.9005(5.0) &  0.9056(3.0) &  0.9210(2.0) &  \bf{0.9229(1.0)} \\
Image                 &  0.4818(2.0) &  0.4809(4.0) &  0.4812(3.0) &  0.4801(5.0) &  \bf{0.4930(1.0)} \\
Langlog               &  0.2401(2.0) & 0.2386(3.0)  &  0.2382(4.0) &  0.0576(5.0) &  \bf{0.2806(1.0)} \\
Medical               &  0.0483(5.0) &  0.4072(2.0) &  \bf{0.4094(1.0)} &  0.0838(3.0) &  0.0645(4.0) \\
Scene                 &  \bf{0.5735(1.5)} &  0.5720(3.0) &  0.5715(4.0) &  0.5416(5.0) &  \bf{0.5735(1.5)} \\
VirusGO               &  0.8365(2.5) &  0.8261(4.0) &  \bf{0.8368(1.0)} &  0.7708(5.0) &  0.8365(2.5) \\
Water-quality         &  \bf{0.5369(1.5)} &  0.5333(3.0) &  0.5329(4.0) &  0.5208(5.0) &  \bf{0.5369(1.5)} \\
Yeast                 &  0.5416(3.0) &  0.5452(2.0) &  \bf{0.5466(1.0)} &  0.5028(5.0) &  0.5400(4.0) \\
\bottomrule
\end{tabular}
        
    \caption{Results for F-Measure for the ECC using all the fusion schemes and data sets.}
    \label{tab:fmeasure}
\end{table}

The results of the F-Measure metric are presented in \refTab{fmeasure}. UDDTECC again proved to be the best both in terms of average ranking and number of best results, outperforming the other methods in 6 of the 15 bases evaluated. MVECC kept the second average ranking with 4 wins and consistent results in the other datasets. Similarly, the DTECC method also obtained 4 best results on data sets. STACKECC was the best in 4 of the evaluated datasets but often showed the worst results, which put it in the last position behind MEECC method.
%, which again, despite inexpressive results, managed to stay among the top 3 in 13 of the 15 evaluated datasets 

\begin{table}[H]
    \centering
\begin{tabular}{llllll}
\toprule
{} &        DTECC &        MEECC &        MVECC &     STACKECC &      UDDTECC \\
Dataset          &              &              &              &              &              \\
\midrule
3sources\_bbc1000      &  0.1133(4.0) &  0.1162(2.5) &  0.1162(2.5) &  0.0996(5.0) &  \bf{0.1786(1.0)} \\
3sources\_guardian1000 &  0.0562(3.5) &  0.0596(2.0) &  0.0562(3.5) &  0.0528(5.0) &  \bf{0.1423(1.0)} \\
3sources\_inter3000    &  0.0000(4.5) &  0.0059(2.5) &  0.0000(4.5) &  0.0059(2.5) &  \bf{0.0180(1.0)} \\
CAL500                &  \bf{0.0000(3.0)} &  \bf{0.0000(3.0)} &  \bf{0.0000(3.0)} &  \bf{0.0000(3.0)} &  \bf{0.0000(3.0)} \\
Emotions              &  0.2327(2.0) &  \bf{0.2361(1.0)} &  0.2310(4.0) &  0.1940(5.0) &  0.2326(3.0) \\
Enron                 &  0.0047(3.5) &  0.0047(3.5) &  0.0053(2.0) &  0.0000(5.0) &  \bf{0.0094(1.0)} \\
Genbase               &  0.0000(5.0) &  0.2764(2.0) &  \bf{0.2779(1.0)} &  0.2388(3.0) &  0.0167(4.0) \\
GnegativeGO           &  \bf{0.8441(2.5)} &  \bf{0.8441(2.5)} &  \bf{0.8441(2.5)} &  0.2443(5.0) &  \bf{0.8441(2.5)} \\
GpositiveGO           &  0.8670(4.5) &  0.8709(2.5) &  0.8670(4.5) &  0.8709(2.5) &  \bf{0.8805(1.0)} \\
Langlog               & 0.1418(4.0) & \bf{0.1641(2.0)} &\bf{0.1641(2.0)} & 0.0042(5.0) & \bf{0.1641(2.0)} \\
Image                 &  0.0740(4.0) &  0.0760(2.0) &  0.0755(3.0) &  \bf{0.0860(1.0)} &  0.0720(5.0) \\
Medical               &  0.0000(4.5) &  0.2710(2.0) &  \bf{0.2730(1.0)} &  0.0010(3.0) &  0.0000(4.5) \\
Scene                 &  \bf{0.1824(1.5)} &  0.1803(3.0) &  0.1778(4.0) &  0.0657(5.0) &  \bf{0.1824(1.5)} \\
VirusGO               &  0.7395(2.5) &  0.7250(4.0) &  \bf{0.7398(1.0)} &  0.4745(5.0) &  0.7395(2.5) \\
Water-quality         &  \bf{0.0019(3.0)} &  \bf{0.0019(3.0)} &  \bf{0.0019(3.0)} &  \bf{0.0019(3.0)} &  \bf{0.0019(3.0)} \\
Yeast                 &  0.1047(4.0) &  \bf{0.1088(1.5)} &  0.1084(3.0) &  0.0732(5.0) &  \bf{0.1088(1.5)} \\
\bottomrule
\end{tabular}
    \caption{Results for Subset Accuracy for the ECC using all the fusion schemes and data sets.}
    \label{tab:subsetaccuracy}
\end{table}

In the Subset Accuracy results presented in \refTab{subsetaccuracy} the UDDTECC method obtained the best result in 10 of the datasets evaluated and stood out as the best in terms of average result.
Although MEECC scored fewer wins than MVECC on the experimental datasets its average ranking was higher. 
MVECC obtained the third best average ranking, doing better than the DTECC method and considerably better than the STACKECC method in most of the datasets evaluated. Observing the relatively low values for this metric one may note the difficulty of optimizing this metric on datasets with large numbers of labels for all the methods evaluated. The CAL500 dataset, which has 174 labels, makes this difficulty evident with all methods obtaining the same value of $0.0$.

\begin{table}[H]
    \centering
\begin{tabular}{llllll}
\toprule
{} &        DTECC &        MEECC &        MVECC &     STACKECC &      UDDTECC \\
Dataset          &              &              &              &              &              \\
\midrule
3sources\_bbc1000      &  0.2174(2.5) &  \bf{0.2155(1.0)} &  0.2179(4.0) &  0.3433(5.0) &  0.2174(2.5) \\
3sources\_guardian1000 &  \bf{0.2037(2.0)} &  \bf{0.2037(2.0)} &  0.2059(4.0) &  0.4411(5.0) &  \bf{0.2037(2.0)} \\
3sources\_inter3000    &  0.2011(3.0) &  \bf{0.2001(1.0)} &  0.2011(3.0) &  0.4643(5.0) &  0.2011(3.0) \\
CAL500                &  0.2909(4.0) &  \bf{0.2855(1.0)} &  0.2892(2.0) &  0.3737(5.0) &  0.2899(3.0) \\
Emotions              &  0.2459(2.0) &  \bf{0.2454(1.0)} &  0.2468(3.5) &  0.2684(5.0) &  0.2468(3.5) \\
Enron                 &  0.1745(2.0) &  0.1797(3.0) &  0.1812(4.0) &  0.6141(5.0) &  \bf{0.1737(1.0)} \\
Genbase               &  0.1615(4.0) &  0.0341(2.0) &  0.0339(1.0) &  0.1251(3.0) &  0.1704(5.0) \\
GnegativeGO           &  \bf{0.0234(2.5)} &  \bf{0.0234(2.5)} &  \bf{0.0234(2.5)} &  0.1405(5.0) &  \bf{0.0234(2.5)} \\
GpositiveGO           &  0.0462(4.5) &  0.0462(4.5) &  0.0453(3.0) &  \bf{0.0443(1.5)} &  \bf{0.0443(1.5)} \\
Langlog               & 0.2066(2.0) & 0.2093(3.0) & 0.2111(4.0) & 0.4124(5.0) & \bf{0.1056(1.0)} \\
Image                 &  0.4115(3.0) &  0.4162(4.0) &  0.4169(5.0) &  \bf{0.3929(1.0)} &  0.4005(2.0) \\
Medical               &  0.5824(5.0) &  \bf{0.0248(1.5)} &  \bf{0.0248(1.5)} &  0.5162(3.0) &  0.5488(4.0) \\
Scene                 &  \bf{0.2324(1.5)} &  0.2337(3.0) &  0.2345(4.0) &  0.2619(5.0) &  \bf{0.2324(1.5)} \\
VirusGO               &  \bf{0.0665(2.0)} &  0.0674(4.0) &  \bf{0.0665(2.0)} &  0.1172(5.0) &  \bf{0.0665(2.0)} \\
Water-quality         &  \bf{0.3998(1.5)} &  0.4233(4.0) &  0.4239(5.0) &  0.4111(3.0) &  \bf{0.3998(1.5)} \\
Yeast                 &  0.3012(3.0) &  \bf{0.2924(1.0)} &  0.2928(2.0) &  0.3615(5.0) &  0.3018(4.0) \\
\bottomrule
\end{tabular}
    \caption{Results for Hamming Loss for the ECC using all the fusion schemes and data sets.}
    \label{tab:hammingloss}
\end{table}

The results presented in \refTab{hammingloss} for the Hamming loss metric differ largely from the others. 
In this metric, unlike the other results, the UDDTECC method ranked second and the MEECC method achieved the best average result. MEECC obtained 8 best results  versus 7 obtained by the UDDTECC method. The DTECC and MVECC methods ranked third and fourth on average, respectively. As with the other metrics, the STACKECC method ranked last in terms of average ranking.

\begin{table}[H]
    \centering
    \caption{
    Average rank of the ECC methods using the evaluated fusion schemes for all datasets.
    }
    \label{tab:avg_all}
\begin{tabular}{lrrrr}
\toprule
{} &  Accuracy & F-measure & Hamming  & Subset \\
Learner  & &  & Loss & Accuracy          \\
\midrule
DTECC & 3.250000 & 3.187500 & 3.500000 & 2.781250 \\
UDDTECC & \bf{1.952381} & \bf{2.125000} & \bf{2.343750} & 2.500000 \\
MEECC & 2.860000 & 3.125000 & 2.437500 & \bf{2.406250} \\
MVECC & 2.600000 & 2.687500 & 2.781250 & 3.156250 \\
STACKECC & 3.565217 & 3.875000 & 3.937500 & 4.156250 \\
\bottomrule
\end{tabular}
\end{table}

\refTab{avg_all} shows the average rank of all methods for each evaluated metric.
The UDDTECC fusion method version was superior on average for the Accuracy, F-measure and Hamming Loss metrics, and it ranks second in Subset Accuracy. The only method other than UDDTECC that achieved the top rank in a metric was the MEECC method in the  Subset Accuracy metric. The MEECC and DTECC methods showed intermediate results with regard to the average ranking of all metrics.
The STACKECC method stood out negatively, presenting the worst average result in all metrics evaluated.
One possible explanation for its poor results is the use of all support values in the classification process performed by the meta-classifier. By using all the support values generated in the classification the meta-classifier uses values that are not correlated to the label being evaluated. These values do not contribute to the result of the method and end up working as noise in the classification process. This being true, it demonstrates the importance of the selection process of correlated labels in UDDTECC since the indiscriminate inclusion of all support values for all labels is actually detrimental to the final result.

\begin{figure}[H]
     \centering
     \begin{subfigure}[b]{0.4\textwidth}
         \centering
         \includegraphics[width=\textwidth]{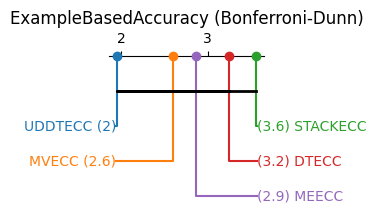}
         \caption{Accuracy}
         \label{fig:cd_acc}
     \end{subfigure}
     \hfill
     \begin{subfigure}[b]{0.4\textwidth}
         \centering
         \includegraphics[width=\textwidth]{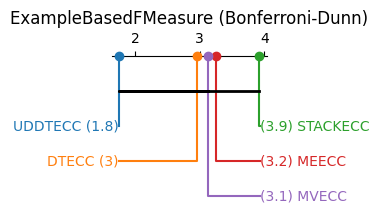}
         \caption{F-Measure}
         \label{fig:cd_fme}
     \end{subfigure}
     \hfill
     \begin{subfigure}[b]{0.4\textwidth}
         \centering
         \includegraphics[width=\textwidth]{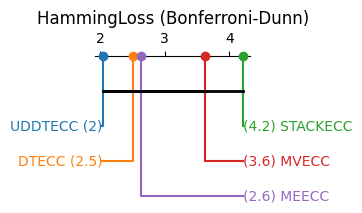}
         \caption{Hamming Loss}
         \label{fig:cd_hml}
     \end{subfigure}
     \hfill
     \begin{subfigure}[b]{0.4\textwidth}
         \centering
         \includegraphics[width=\textwidth]{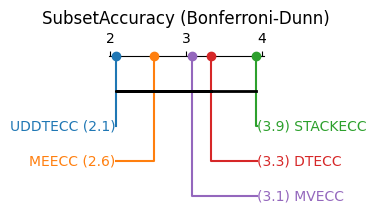}
         \caption{Subset Accuracy}
         \label{fig:cd_sac}
     \end{subfigure}
     \hfill
        \caption{Critical difference (CD) diagram comparing all methods across evaluation metrics.}
        \label{fig:cd_all_datasets}
\end{figure}

Statistical analysis used the Bonferroni-Dunn test, selected for its rigorous error control in multiple comparisons and suitability to evaluate UDDTECC against baselines in our datasets. Although conservative, this test provides strong protection against false positives, complementing numerical ranking analysis \cite{demvsar2006statistical}. 

First, we applied the Friedman test to determine whether statistically significant differences existed among the methods (null hypothesis: all methods perform equivalently). The test rejected the null hypothesis ($p < 0.1$), indicating significant differences between at least some of the compared methods. The Bonferroni-Dunn test ($\alpha=0.1$, adjusted for $k-1$ comparisons) did not reveal statistically significant differences, a frequent outcome with moderate sample sizes (16 datasets) due to reduced statistical power. 

Nevertheless, UDDTECC's consistent performance advantage proves particularly noteworthy in high-diversity scenarios where label correlations exhibit greater complexity, as we will analyze in the following subsection.

\subsection{Higher Diversity Datasets Results}
\label{sec:HDDR}
Diversity indicates the fraction of label sets present in a dataset relative to the maximum possible number of sets. Thus, lower diversity values indicate that a dataset has more predictable label sets than datasets with high diversity values. 
When one looks specifically at the experimental results for datasets with higher diversity ($Diversity > 0.1$) one may see that the best UDDTECC results are concentrated exactly in datasets with this characteristic.  \refTab{avg_div} shows the average results of the analyzed metrics  focusing on the 12 datasets (3sources\_bbc1000,
3sources\_guardian1000,
3sources\_inter3000,
CAL500,
Emotions,
Enron,
GpositiveGO,
Langlog,
Image,
Scene,
VirusGO,
Water-quality) whose Diversity $> 0.1$.

\begin{table}[ht]
    \centering
    \caption{
    Average rank of the ECC methods using the evaluated fusion schemes for higher diversity datasets.
    }
    \label{tab:avg_div}
\begin{tabular}{lrrrr}
\toprule
{} &  Accuracy & F-measure & Hamming  & Subset \\
Learner  & &  & Loss & Accuracy          \\
\midrule
DTECC & 3.250000 & 2.958333 & 2.500000 & 3.333333  \\
UDDTECC & \bf{1.952381} & \bf{1.750000} & \bf{2.041667} & \bf{2.083333}  \\
MEECC & 2.860000 & 3.250000 & 2.625000 & 2.583333  \\
MVECC & 2.600000 & 3.125000 & 3.625000 & 3.083333  \\
STACKECC & 3.565217 & 3.916667 & 4.208333 & 3.916667  \\
\bottomrule
\end{tabular}
\end{table}

The experimental results \reftab{avg_div} demonstrate the superior performance of UDDTECC across all metrics in data sets with relatively high label diversity, achieving the top rankings in accuracy (1.95), F-measure (1.75), Hamming Loss (2.04), and Subset Accuracy (2.08).  

\begin{figure}[H]
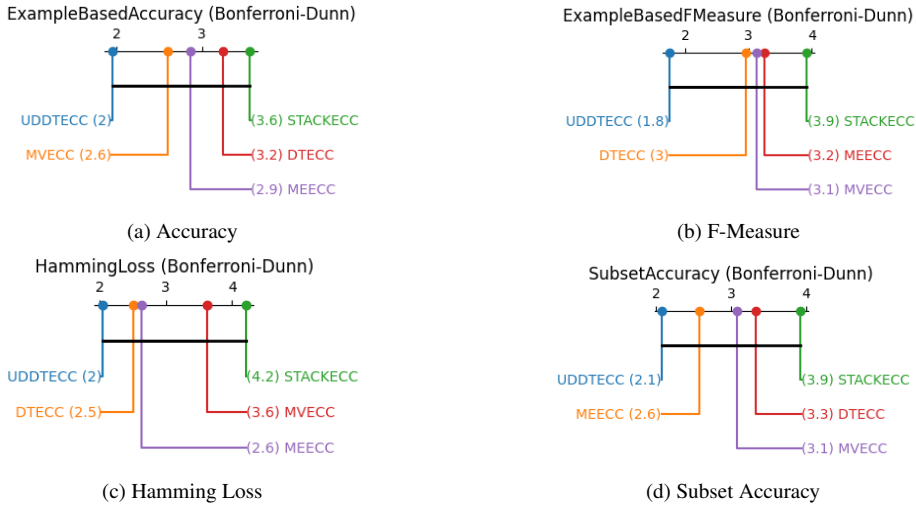

     \centering
     \begin{subfigure}[b]{0.4\textwidth}
         \centering
         \includegraphics[width=\textwidth]{bd_cd_acc_a010.png}
         \caption{Accuracy}
         \label{fig:cd_acc}
     \end{subfigure}
     \hfill
     \begin{subfigure}[b]{0.4\textwidth}
         \centering
         \includegraphics[width=\textwidth]{bd_cd_fme_a010.png}
         \caption{F-Measure}
         \label{fig:cd_fme}
     \end{subfigure}
     \hfill
     \begin{subfigure}[b]{0.4\textwidth}
         \centering
         \includegraphics[width=\textwidth]{bd_cd_hml_a010.png}
         \caption{Hamming Loss}
         \label{fig:cd_hml}
     \end{subfigure}
     \hfill
     \begin{subfigure}[b]{0.4\textwidth}
         \centering
         \includegraphics[width=\textwidth]{bd_cd_sac_a010.png}
         \caption{Subset Accuracy}
         \label{fig:cd_sac}
     \end{subfigure}
     \hfill
        \caption{Critical difference (CD) diagrams for comparing the average of the methods evaluated in the 12 data sets with the highest diversity. The diagram shows average ranks (left axis) with connecting lines indicating groups of methods not significantly different at $\alpha=0.1$ (Bonferroni-Dunn test). Lower ranks indicate better performance. The critical difference threshold (right axis) determines the minimum rank difference required for statistical significance.}
        \label{fig:cd_high_diversity}
\end{figure}

Both \reffig{cd_high_diversity} and \reffig{cd_all_datasets} show connected critical difference intervals, indicating that no statistically significant differences were found between the compared methods in either scenario, including the high-diversity datasets. Although this outcome is not favorable to the proposed UDDTECC, it reflects the challenging nature of multi-label classification when dealing with complex label dependencies. Nevertheless, the practical significance of UDDTECC's rankings suggests its dynamic threshold adaptation offers meaningful benefits in such contexts, especially where traditional methods struggle with label interdependence.

\subsection{Label Correlation Analysis}
\label{sec:correlation}

The proposed UDDTECC method introduces a novel approach to label correlation that differs fundamentally from ECC's chain rule mechanism. While ECC models correlations sequentially through classifier chains (where prediction of $y_j$ depends on $y_1,...,y_{j-1}$), UDDTECC captures correlations through dynamic threshold adaptation based on pairwise label dependencies. This distinction creates two complementary layers of correlation modeling: ECC at the base classifier level and UDDTECC at the decision fusion level.

To empirically validate this complementarity, this work considers as the baseline the Binary Relevance (BR) multi-label classifier \cite{zhang2018br}, which inherently disregards label correlation, to see whether and how much the proposed approach can improve its performance. thus, experiments were conducted using an Ensemble of Binary Relevance (EBR) baseline. By applying UDDT to EBR (yielding UDDTEBR), this work isolates the effects of UDDT's correlation handling. \refTab{ebr_results} shows consistent improvements in all metrics, with UDDTEBR achieving better rankings in F-measure (1.15 tvs 1.85), Accuracy (1.25 vs 1.75), Subset Accuracy (1.35 vs 1.65) and Hamming Loss (1.35 vs 1.65).

\begin{table}[H]
\centering
\caption{Performance comparison between EBR and UDDTEBR}
\label{tab:ebr_results}
\begin{tabular}{lcc}
\hline
\textbf{Metric} & \textbf{EBR} & \textbf{UDDTEBR} \\ \hline
F-measure & 1.85 & 1.15 \\
Accuracy & 1.75 & 1.25 \\
Subset Accuracy & 1.65 & 1.35 \\
Hamming Loss & 1.65 & 1.35 \\ \hline
\end{tabular}
\end{table}

These findings indicate that UDDT's correlation modeling remains effective even when applied to base methods that do not take correlations into account and no conflict occurs between UDDT's approach and ECC's chain rule mechanism. Therefore, the methods operate at different abstraction levels, suggesting potential synergistic effects when combined. 

\section{Conclusions} \label{sec:conclusion}

This paper presented the UDDTECC method. A trainable multi-label classifier fusion method based on the single-label classifier combination scheme called Decision Templates. The main features that distinguish the UDDTECC method from other fusion methods typically used in multi label classification are its ability to adapt to the problem it is being used on and the exploitation of unconditional dependencies between problem labels. Methods commonly used in Multi-Label ensembles, such as MV and ME, consist of applying fixed and invariant rules regardless of the problem to which they are being applied.

In this study, the UDDTECC method is evaluated using a cross-validation scheme in which it is directly compared to the most commonly used fusion schemes in multi-label classifier fusion: Majority Vote and MeanEnsemble. STACKECC, another trainable fusion model is also included in the tests to see how the UDDTECC would perform against a stacking method given the similarity between the two strategies. The DTECC method on which the UDDTECC is based was also included in the experiments.

Of the four metrics evaluated, the UDDTECC method was the best on average in three of them: Accuracy, F-measure and Subset-Accuracy. In Hamming Loss, the method was in second place, losing to MEECC in terms of average results. The STACKECC method, on the other hand, despite obtaining competitive results in very few datasets, obtained the worst average result in all the metrics evaluated, showing that the strategy of using trainable functions is not necessarily superior to fixed combination rules in all cases. The use of all unfiltered information by the STACKECC method may have introduced noise (information about labels unrelated to what is being evaluated) which would explain the low performance of the method.
Using label information that does not correlate to what is being evaluated can be detrimental in the classification process. In the way it was proposed, the meta-classifier is responsible for filtering the useful attributes and performing the classification process, a strategy that did not prove to be good enough in the experiments. 

Future work should extend the performance evaluation of the UDDTECC fusion schemes to different ensemble-based multi-label classification methods other than ECC. Using the UDDTECC method with different classification methods, such as \RAKEL~\cite{tsoumakas2007random}, Ensemble of Subset Learners \cite{tenenboim2010identification} and Ensemble of Pruned Sets \cite{read2008multi}, can give us valuable insight into the potential of its use. However, adapting UDDT for \RAKEL would require firstly redesigning confidence combination through approaches like partial subset correlation aggregation, and secondly reengineering the rank fusion mechanics. Given these complexities, future work should prioritize methods like ECC where UDDT can be applied directly and its benefits are clearly measurable.

\section*{Declarations}

\begin{itemize}
    \item[$\bullet$] \textbf{Funding}
    \item[] Partial financial support was received from Conselho Nacional de Desenvolvimento Científico e Tecnológico (CNPq, Brazil).
    \item[$\bullet$] \textbf{Conflicts of interest/Competing interests}
    \item[] The authors declare that they have no conflict of interest.
    \item[$\bullet$] \textbf{Ethics approval}
    \item[] Not applicable.
    \item[$\bullet$] \textbf{Consent to participate}
    \item[] Not applicable.
    \item[$\bullet$] \textbf{Consent for publication}
    \item[] Not applicable.
    \item[$\bullet$] \textbf{Availability of data and material}
    \item[] The data that support the findings of this study are available in Multi-Label Classification Dataset Repository at \href{https://www.uco.es/kdis/mllresources/}{https://www.uco.es/kdis/mllresources/}.
    \item[$\bullet$] \textbf{Code availability}
    \item[] The custom code used in this article was made available in a public repository at \\ \href{https://github.com/vfrocha/ddtecc}{https://github.com/vfrocha/ddtecc}.
    \item[$\bullet$] \textbf{Authors' contributions}
    \item[] The study's conception and design involved contributions from all authors. Alexandre L. Rodrigues, Thiago Oliveira-Santos, and Flávio M. Varejão conducted the analysis of the results. The first draft of the manuscript was written by Victor F. Rocha and all authors commented on previous versions of the manuscript. All authors read and approved the final manuscript.
\end{itemize} 

\bibliographystyle{vancouver}
\bibliography{paper}
\end{document}